%% file: main.tex
\RequirePackage{caption}
\let\abovecaptionskip\undefined
\let\belowcaptionskip\undefined
\documentclass[lettersize,journal]{IEEEtran}

\newif\ifsupp
\supptrue

\PassOptionsToPackage{svgnames,dvipsnames}{xcolor}
\PassOptionsToPackage{capitalize}{cleveref}
\PassOptionsToPackage{pdftex}{graphicx}

\usepackage{amsmath,amssymb}
\usepackage{duckuments}
\usepackage{siunitx}
\usepackage{makecell}
\usepackage{tabularray}
\UseTblrLibrary{booktabs,siunitx}
\usepackage{makecell}
\usepackage{ifthen}
\usepackage{bm}
\usepackage{float}
\usepackage{enumitem}
\usepackage{physics}
\usepackage{xspace}
\usepackage{color}
\usepackage{xcolor}
\usepackage{mathtools}

\DeclareSIUnit{\GB}{\giga\byte}
\DeclareSIUnit[quantity-product=]{\%}{\percent}
\DeclareMathOperator*{\argmin}{argmin}

\newcommand{\sitime}[2]{#1'~#2''}

\NewTableCommand{\BB}{\SetCell{font=\bfseries\sffamily}}

\makeatletter
\renewcommand\paragraph{\@startsection{paragraph}{4}{\z@}%
 {-0.8ex \@plus -0.3ex \@minus -0.3ex}%
 {-0.0em \@plus -0.2em \@minus -0.2em}%
 {\normalfont\normalsize\bfseries}}
\makeatother
\let\oldparagraph\paragraph
\renewcommand\paragraph[1]{\oldparagraph*{#1}}

\usepackage[T1]{fontenc}
\DeclareSymbolFont{Letters}{U}{zeur}{m}{n} 
\DeclareMathSymbol{\epsilon}{\mathalpha}{Letters}{"22}

\newcommand{\insertfig}{
 \includegraphics[width=\linewidth]{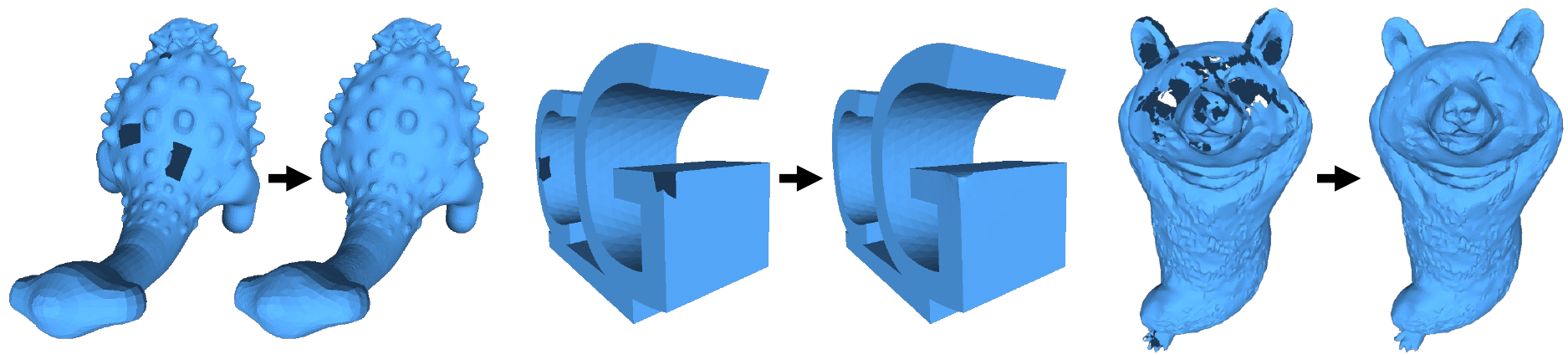}
 \captionof{figure}{Our self-prior-based method inpaints holes or missing regions in triangular meshes through self-supervised learning from a single incomplete mesh, without using any training data. As demonstrated in this figure, our method successfully fills the holes over the characteristic bumps on a dinosaur's back (left), the sharp edges of an extruded text ``CG'' (middle), and the complex missing regions inevitably obtained from a real scan (right).}
}
\makeatletter
\apptocmd{\@maketitle}{%
 \setcounter{figure}{0}%
 \centering\insertfig}{}{}
\makeatother

\usepackage[
 bibstyle=ieee,
 citestyle=numeric,
 defernumbers=false,
 bibencoding=utf-8,
 sortcites=true,
 backend=biber,
 natbib=true,
 doi=true,
 url=true,
 isbn=false,
]{biblatex}
\let\oldmulticitedelim\multicitedelim
\renewcommand{\multicitedelim}{\bibclosebracket\oldmulticitedelim\bibopenbracket}

\usepackage{confnames}

\usepackage[pdftex]{graphicx}
\graphicspath{{./figure/}}

\usepackage{amsmath,amssymb}

\usepackage{algorithmic}

\usepackage{array}

\usepackage[caption=false,font=footnotesize,labelfont=sf,textfont=sf]{subfig}

\usepackage{stfloats}

\ifCLASSOPTIONcaptionsoff
 \usepackage[nomarkers]{endfloat}
\let\MYoriglatexcaption\caption
\renewcommand{\caption}[2][\relax]{\MYoriglatexcaption[#2]{#2}}
\fi

\usepackage{hyperref}
\hypersetup{
 pdftex,
 pdftitle={Learning Self-Prior for Mesh Inpainting Using Self-Supervised Graph Convolutional Networks},
 pdfauthor={Shota Hattori, Tatsuya Yatagawa, Yutaka Ohtake, Hiromasa Suzuki},
 bookmarks=true,
 bookmarksnumbered=false,
 pdfstartview=FitH,
 colorlinks=true,
 urlcolor=magenta,
 citecolor=Blue,
 linkcolor=Blue,
}
\usepackage[capitalize]{cleveref}

\begin{document}
%
\title{Learning Self-Prior for Mesh Inpainting Using Self-Supervised Graph Convolutional Networks}
%
%
%
%

\author{Shota~Hattori, Tatsuya~Yatagawa, \IEEEmembership{Member,~IEEE}, Yutaka~Ohtake, and~Hiromasa~Suzuki
  \IEEEcompsocitemizethanks{
    \IEEEcompsocthanksitem
    S. Hattori, Y. Ohtake, and H. Suzuki are with the School of Engineering, the University of Tokyo, Tokyo, Japan, 113-8656\protect\\
    E-mail: shota.hattori.den@gmail.com (S. Hattori), ohtake@den.t.u-tokyo.ac.jp (Y. Ohtake), suzuki@cpe.or.jp (H. Suzuki)
    \IEEEcompsocthanksitem
    T. Yatagawa is with the School of Social Data Science, Hitotsubashi University, Tokyo, Japan, 186-8601.\protect\\
    E-mail: tatsuya.yatagawa@r.hit-u.ac.jp}
}

%
%

\markboth{Hattori \MakeLowercase{\textit{et al.}}: Learning Self-Prior for Mesh Inpainting Using Self-Supervised Graph Convolutional Networks}%
{Hattori \MakeLowercase{\textit{et al.}}: Learning Self-Prior for Mesh Inpainting Using Self-Supervised Graph Convolutional Networks}
%



\IEEEtitleabstractindextext{%
  \begin{abstract}
    In this paper, we present a self-prior-based mesh inpainting framework that requires only an incomplete mesh as input, without the need for any training datasets. Additionally, our method maintains the polygonal mesh format throughout the inpainting process without converting the shape format to an intermediate one, such as a voxel grid, a point cloud, or an implicit function, which are typically considered easier for deep neural networks to process. To achieve this goal, we introduce two graph convolutional networks (GCNs): single-resolution GCN (SGCN) and multi-resolution GCN (MGCN), both trained in a self-supervised manner. Our approach refines a watertight mesh obtained from the initial hole filling to generate a complete output mesh. Specifically, we train the GCNs to deform an oversmoothed version of the input mesh into the expected complete shape. The deformation is described by vertex displacements, and the GCNs are supervised to obtain accurate displacements at vertices in real holes. To this end, we specify several connected regions of the mesh as fake holes, thereby generating meshes with various sets of fake holes. The correct displacements of vertices are known in these fake holes, thus enabling training GCNs with loss functions that assess the accuracy of vertex displacements. We demonstrate that our method outperforms traditional dataset-independent approaches and exhibits greater robustness compared with other deep-learning-based methods for shapes that infrequently appear in shape datasets. Our code and test data are available at {\url{https://github.com/astaka-pe/SeMIGCN}}.
  \end{abstract}

  }

\maketitle

\thispagestyle{plain}

\IEEEdisplaynontitleabstractindextext

%
\IEEEpeerreviewmaketitle


%
%
%
%

\section{Introduction}
\label{sec:introduction}

\IEEEPARstart{T}{he} triangular mesh is one of the most common representations of object geometries as digital data. Recently, three-dimensional (3D) scanning applications of real-world objects have rapidly been spreading owing to the prevalence of low-cost 3D scanning devices. However, mesh data captured by these 3D scanners often include holes or missing regions because of self-occlusion, dark colors, and high surface specularity. Therefore, filling these holes, which is technically referred to as ``inpainting,'' is an essential process in digitizing real-world objects and using the data not only in graphics applications, such as digital modeling, rendering synthetic 3D scenes, and 3D printing, but also in industrial applications, such as simulation-based evaluation and nondestructive inspection of real products.

Inpainting is an ill-posed problem that involves restoring various types of incomplete media, such as images and polygonal meshes, where the desired output is not uniquely determined from the available data. Thus, some prior knowledge must be introduced to solve this problem. In mesh inpainting, heuristic priors, such as smoothness and self-similarity of the input geometry, have been used traditionally as prior knowledge~\cite{barequet1995minarea,liepa2003minmax,clarenz2004finite,pernot2006filling,zhao2007robust,Daniel2004context,harary2014coherent}. Although these heuristic priors often work well for simple small holes, they require careful parameter tuning and user intervention to obtain good results.

In contrast to the prior knowledge based on human heuristics, data-driven priors, i.e., those acquired from datasets by deep learning, have recently been leveraged in inpainting images, as presented in comprehensive surveys~\cite{jam2021comprehensive,xiang2023deep}. Data-driven priors have also been used for several types of 3D geometries, such as voxel grids~\cite{wu20153d} and point clouds~\cite{dai20173depn,yuan2018pcn,wen2020point}. However, to our knowledge, deep-learning-based mesh inpainting, particularly methods that do not convert the input mesh to another shape format, has not yet been sufficiently explored. The main challenge lies in the need to add triangles to the missing regions of the input mesh while processing it using a deep neural network. Unfortunately, even with state-of-the-art networks~\cite{feng2019meshnet,hanocka2019meshcnn,rakotosaona2021differentiable,sharp2022diffusionnet}, dynamic triangulation involving changes in vertex connections remains challenging. This limitation hinders the development of deep-learning-based mesh inpainting methods that do not rely on an intermediate format, despite the existence of several large-scale mesh datasets~\cite{Thingi10K,koch2019abc}.

Recently, the ``self-prior,'' i.e., prior knowledge acquired from only a single damaged input to solve an ill-posed problem, has attracted attention in the image processing field. Deep Image Prior (DIP)~\cite{ulyanov2018dip} is a seminal study that acquires the self-prior from a single input and performs various image restoration tasks. However, it was also shown that inpainting with DIP is more challenging than other restoration tasks, such as denoising and super-resolution, due to the lack of supervision in the missing regions. Although there are several follow-up studies to solve several problems~\cite{mataev2019deepred,cheng2019bayesian,jo2021rethinking} of the original DIP, most focus on image denoising, and thus, inpainting by DIP and its variant is still challenging even in the image processing field.

Self-prior-based methods for reconstructing surface geometries from point clouds~\cite{hanocka2020point2mesh,wei2021deephybrid} can be used for mesh inpainting by either treating the vertices of the input mesh as a point cloud or sampling points on the surface. Although these methods do not require large shape datasets and work with only a single input, they might change the topology, such as the genus, of the input surface geometry to an inappropriate one by processing the mesh as a point cloud. Similarly, several research groups have explored self-supervised and unsupervised learning for shape inpainting~\cite{chu2021unsupervised,mittal2021selfpcinpaint}. While these methods do not need ground-truth inpainted geometries, they still require large shape datasets. Moreover, these methods involve converting input meshes to other shape formats, which might still unpredictably change the topological properties of an input shape.

To address the above problems, we present a self-supervised learning method for mesh inpainting that retains the mesh format throughout the process, leveraging the self-prior and requiring only a single mesh with missing regions as input. Our method offers three key advantages over previous methods. First, it eliminates the need for large-scale datasets, as it is based on the self-prior and relies on a single damaged mesh. Second, self-prior-based methods are inherently approaches that only use the input media, which allows us to use a graph convolutional network (GCN), typically when working on static graph structures, by defining it on the graph structure of the input mesh. We introduce two types of GCNs: the single-resolution graph convolutional network (SGCN) and the multi-resolution graph convolutional network (MGCN). SGCN consists of straightly stacked graph convolutional layers on the same graph structure as the input mesh, while MGCN is an hourglass-type encoder--decoder network with pooling and unpooling layers, defined using precomputed progressive meshes~\cite{hoppe1996progressive} for the input mesh. Third, our method is also a self-supervised learning method that performs data augmentation on the single input mesh by marking random parts as fake holes, where the correct vertex positions are available for supervision during training.

\paragraph{Contributions} In summary, the contributions of this study include the following.
\begin{itemize}
  \item Our method directly inpaints triangular meshes, eliminating the need for conversion to other formats, such as point clouds, voxel grids, and implicit functions.
  \item Our self-prior-based method operates without shape datasets and inpaints the incomplete input mesh on the basis of self-supervised learning. Thus, like traditional methods, it works solely with a single input mesh.
  \item We investigate the inpainting performance of our two proposed network architectures, SGCN and MGCN, demonstrating their state-of-the-art performance on various triangular meshes.
\end{itemize}
To our knowledge, we are the first to use the self-prior for mesh inpainting. However, as shown in the original DIP paper~\cite{ulyanov2018dip}, inpainting with the self-prior is more challenging than other tasks, such as denoising. Therefore, rather than proposing the single best approach, we investigate the effects of data augmentation for self-supervision, network architectures, and loss functions on the final output mesh.




\section{Related Work}
\label{sec:related-work}

\subsection{Traditional Mesh Inpainting}

Traditional mesh inpainting methods based on geometry processing are roughly classified into surface-based and volume-based methods.

Surface-based methods first identify boundaries and then interpolate holes directly by generating triangles based on local geometric properties. These methods typically guarantee the quality of the generated triangles by applying remeshing or smoothing to optimize vertex density, triangle area, or dihedral angles~\cite{barequet1995minarea,liepa2003minmax,clarenz2004finite,pernot2006filling,zhao2007robust}. These methods work well for completing simple small holes, but are prone to fail for large complicated holes. Unlike these methods that rely solely on local geometric processing, context-based inpainting methods have also been proposed to enhance self-similarity by synthesizing features of similar patches explored in the input~\cite{Daniel2004context,harary2014coherent}. Additionally, Kraevoy and Sheffer~\cite{kraevoy2005tempcomp} proposed a template-based inpainting method that utilizes mapping between the incomplete input mesh and a template model. The benefits of those context-based and template-based methods lie in their capability to restore detailed geometric features, whereas their drawbacks include the dependence on the presence of similar geometric features, complete template meshes, and user interactions.

Volume-based methods initially convert an input mesh into a discrete voxel representation and then apply several inpainting methods. For example, volumetric diffusion is a typical approach to obtain complete voxel representations~\cite{davis2002diffcomp,verdera2003inpainting,guo2006filling}. As with these studies, Bischoff~et~al.~\cite{bischoff2005automatic} proposed morphological dilation and erosion for hole filling on a time-efficient octree structure. After filling holes in the voxel representation, a final mesh is obtained by generating triangles on the isosurface. These volume-based methods are beneficial because they can fill holes in complicated surface geometry and generate manifold meshes. The drawback is that the shape of the remaining areas also deforms during the diffusion and morphing phase. Moreover, the output resolution of these methods is often limited because of the computational inefficiency of the voxel representation.

\subsection{Data-Driven Shape Inpainting}

Advances in deep learning have led to increased interest in data-driven inpainting methods. For voxel grids and point clouds, general-purpose deep neural networks, such as 3D convolutional neural network (CNN)~\cite{lecun19983dcnn}, PointNet~\cite{qi2017pointnet}, and PointNet++~\cite{qi2017pointnetpp}, have facilitated the development of shape inpainting approaches.

For shape completion on voxel grids, various 3D CNNs have been proposed. Wu~et~al.~\cite{wu20153d} proposed volume-based shape completion from depth maps using a convolutional deep belief network (CDBN). Varley~et~al.~\cite{Varley2017robot} proposed shape completion for robotic grasping. However, the high computational cost of these methods hinders the processing of high-resolution voxel grids. Subsequently, Dai~et~al.~\cite{dai20173depn} proposed a 3D encoder--predictor network (3D-EPN) for high-resolution shape completion by combining coarse shape prediction and multi-resolution shape synthesis. Han~et~al.~\cite{han2017high} achieved a high-resolution shape completion by jointly training two networks, one inferring global structure and the other focusing on local geometry.

Point cloud completion has also been studied because the point cloud is a direct output of many 3D scanning devices. The point completion network (PCN)~\cite{yuan2018pcn} and skip-attention network (SA-Net)~\cite{wen2020point} first encode an incomplete point cloud into the global feature and then decode the feature to a complete point cloud. Generative adversarial networks (GANs) have also been used for point cloud completion. Chen~et~al.~\cite{chen2020unpaired} proposed GAN-based point cloud completion for real scans that does not require a training dataset consisting of pairs of incomplete and complete point clouds. Sarmad~et~al.~\cite{sarmad2019rl} proposed a reinforcement learning method using a GAN for point cloud completion. PMP-Net~\cite{wen2021pmp}, a neural network inferring point moving paths (PMP), restores the complete point cloud by learning the movement from the initial input, rather than generating new points directly.

Recently, shape inpainting using implicit functions has attracted increasing attention because it can be easily represented by a neural network in a memory-efficient manner. For example, several approaches learn a signed distance function (SDF) that represents the distance from each point of an input point cloud to its underlying surface geometry~\cite{chibane2020implicit,park2019deepsdf,rao2022patchcomplete}. In addition, other approaches using the vector quantized deep implicit function (VQDIF)~\cite{yan2022shapeformer,mittal2022autosdf}, a more computationally efficient implicit representation using a deep neural network, have also been proposed.

Although there are many studies on shape inpainting using voxel grids, point clouds, and implicit functions, few studies have been conducted for mesh inpainting because general-purpose deep neural networks for meshes have not been well established despite the recent efforts~\cite{feng2019meshnet, hanocka2019meshcnn,rakotosaona2021differentiable, sharp2022diffusionnet}. In addition, all the methods described above require long pretraining times with large-scale datasets.

Mittal et al.~\cite{mittal2021selfpcinpaint} and Chu et al.~\cite{chu2021unsupervised} both proposed methods for shape inpainting based on self-supervised and unsupervised learning. Mittal et al. focused on inpainting point clouds using a self-supervised deep learning approach, which assumes virtual occlusion from the optical range scanner and trains a neural network to reproduce the positions of removed points. However, applying this method to polygonal meshes requires converting them to point clouds, which may lose the topological properties of the input surface. Chu et al.'s method processes incomplete polygonal meshes using a truncated signed distance field (TSDF) for inpainting missing regions of real 3D scans. This approach allows for flexible changes in topological properties owing to TSDF; however, as with Mittal et al.'s method, this flexibility can be a disadvantage in the sense that it may change the topological properties of the input surface geometry to an inappropriate one. Moreover, while both methods share the advantage of not requiring complete ground-truth shapes, they still require large-scale training datasets, such as ShapeNet~\cite{chang2015shapenet}, KITTI~\cite{behley2019semantickitti}, and Thingi10k~\cite{Thingi10K}. Consequently, they may fail in inpainting geometries that are rarely included in such datasets.

In contrast, our method relies on only a single incomplete mesh and preserves the topological properties of input surface geometry without converting the mesh to another format. Thus, our method eliminates the need for shape datasets while maintaining the original topological properties, providing a more reliable solution for mesh inpainting.

\subsection{Self-Prior in Geometry Processing}

Recently, the ``self-prior,'' the prior knowledge acquired from a single damaged input, has attracted much attention. DIP~\cite{ulyanov2018dip}, i.e., the seminal study in the image processing field, can be applied to various image restoration tasks, such as denoising, super-resolution, and inpainting. The original DIP had several problems, such as instability in optimizing network parameters, unclear termination criterion of the optimization, and long processing time. These problems have been solved over time. For example, the instability and termination criterion issues have been addressed using a denoising regularizer~\cite{mataev2019deepred}, a Bayesian formulation~\cite{cheng2019bayesian}, and Stein's unbiased risk estimator~\cite{jo2021rethinking}; the long processing time has been addressed by applying meta learning~\cite{zhang2022metadip}.

The self-prior has also been applied in the geometry processing field. The deep geometric prior (DGP)~\cite{williams2019dgp} acquires a self-prior for surface reconstruction from a single input point cloud by training a separate multilayer perceptron (MLP) for each local region. Point2Mesh~\cite{hanocka2020point2mesh} is similar to DGP but uses MeshCNN~\cite{hanocka2019meshcnn}, which consists of weight-sharing convolution layers and dynamic mesh pooling/unpooling layers. Point2Mesh was later extended for textured mesh generation from colored point clouds~\cite{wei2021deephybrid}. For sparse point clouds, the problem of shape reconstruction, which the aforementioned methods addressed, becomes more challenging. To compensate for the sparsity of input points, recent studies have proposed various techniques, such as leveraging the SDF format for unsupervised mesh reconstruction~\cite{ma2022reconstructing,chen2023unsupervised} and resolving the irregularity of point density in a point cloud by the technique known as point consolidation~\cite{metzer2021self}. However, even though these methods address the sparsity of point clouds in an unsupervised manner, mesh inpainting is still an unaddressed problem. For techniques for meshes, Hattori~et~al.~\cite{hattori2022ddmp} applied the self-prior to mesh restoration but only focused on mesh denoising. Thus, the self-prior for mesh inpainting has not been sufficiently investigated.

\begin{figure*}[t!]
  \centering
  \includegraphics[width=\linewidth]{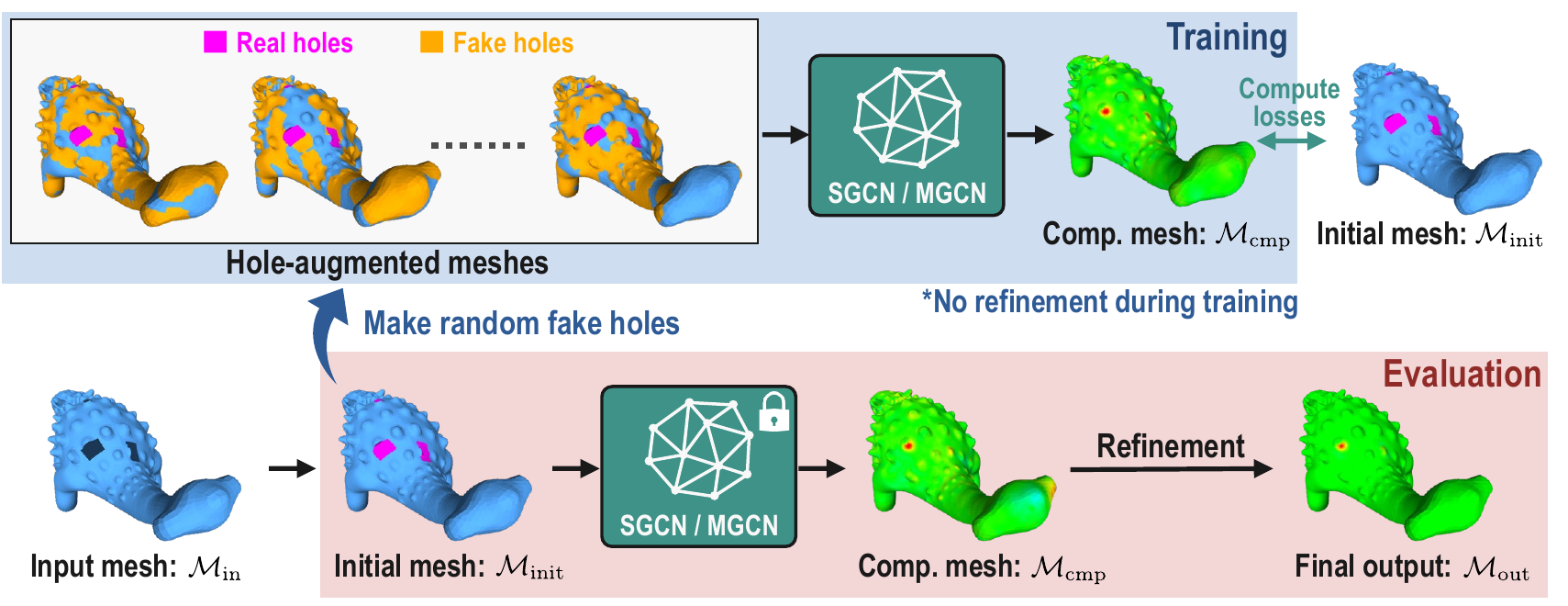}
  \caption{Overview of our self-supervised mesh inpainting. Our method, using only an input mesh with missing regions, consists of training and evaluation phases. In the training phase, the GCN is trained with hole-augmented meshes having different fake hole regions. The evaluation phase inputs the mesh $\mathcal{M}_{\text{in}}$ with only real holes and outputs the hole-completed mesh $\mathcal{M}_{\text{cmp}}$. Refinement, performed only in the evaluation phase, is then applied to improve the vertex positions at both hole and non-hole regions.}
  \label{fig:overview}
\end{figure*}

\section{Self-Supervised Mesh Inpainting by GCNs}
\label{sec:method}

\Cref{fig:overview} shows the overview of our method, which consists of training and evaluation phases illustrated at the top and bottom, respectively. As explained later, our method solves the inpainting problem by moving the vertices of a smoothed input mesh to their correct positions. In other words, we ``inpaint'' the displacement vectors for the vertices rather than inpainting the positions of the vertices. Thus, we first fill the holes of the input mesh by a conventional hole-filling method.

During the training phase, we train a GCN using data augmentation, where several continuous regions on a single input mesh are marked as fake holes. We input a 4D vector into the GCN, where the first three entries represent a 3D vertex displacement, and the last entry is either 0, indicating the presence of a real/fake hole, or 1, indicating no hole, to represent whether the corresponding vertex is masked by a hole. If a vertex is masked, the 3D vertex displacement part is also filled with zeros. This enforces the GCN to treat displacement vectors at fake holes as unknown, just like those at real holes. In this manner, we train the GCN to predict displacement vectors in hole regions, allowing for self-supervised learning, because we know the correct displacement vectors at fake holes.

During the evaluation phase, only the vertex displacements at real holes are estimated by the GCN without using fake holes. We obtain the complete mesh by applying the displacements to the oversmoothed mesh. To further refine the precise shape of non-hole regions, we correct the vertex positions of both hole and non-hole regions by solving a least-squares problem~\cite{lipman2004differential}.

\paragraph{Notations}

We denote a mesh with the symbol $\mathcal{M}$, which may have boundaries and nonzero genus, and denote the sets of vertices and faces of $\mathcal{M}$ as $\mathcal{V}$ and $\mathcal{F}$, respectively. The positions of vertices and the normal vectors of faces are denoted as $\vb{X} \in \mathbb{R}^{|\mathcal{V}| \times 3}$ and $\vb{N} \in \mathbb{R}^{|\mathcal{F}| \times 3}$, respectively, where $|\mathcal{S}|$ is the size of the set $\mathcal{S}$. Additionally, we denote the $i$-th vertex position as $\vb{x}^i \in \mathbb{R}^3$ and the $j$-th facet normal as $\vb{n}^j \in \mathbb{S}^2$ (i.e., the set of unit vectors in $\mathbb{R}^3$). Furthermore, we denote the input mesh as $\mathcal{M}_{\text{in}}$ and refer to intermediate output forms of the mesh by different names. The initial mesh $\mathcal{M}_{\text{init}}$ refers to the mesh after preprocessing (see \cref{ssec:preprocess}), which is the input for the proposed networks described in \cref{ssec:network}. As described in a later section, $\mathcal{M}_{\text{init}}$ is a watertight mesh, which ensures the manifold property of the final output mesh. The surface of $\mathcal{M}_{\text{init}}$ is classified into the source region, which exists in the input mesh, and the target region, which is inserted by initial hole filling. We refer to the output mesh from the networks as a complete mesh $\mathcal{M}_{\text{cmp}}$ with revised vertex positions at the target region. Finally, we revise the vertex positions of $\mathcal{M}_{\text{cmp}}$ by the refinement step (see \cref{ssec:refinement}) to obtain the final output mesh $\mathcal{M}_{\text{out}}$. We refer to the holes of the input mesh and those made during data augmentation (see \cref{ssec:augmentation}) as ``real'' holes and ``fake'' holes, respectively. To identify vertices and normals in these hole regions, we introduce masks: $m_{\text{vtx}}^i$ for the $i$-th vertex, and $m_{\text{nrm}}^j$ for the $j$-th facet normal. These masks take a value of $0$ in hole regions and $1$ otherwise.

\subsection{Input Mesh Preprocessing}
\label{ssec:preprocess}

\begin{figure}[t!]
  \centering
  \includegraphics[width=0.9\linewidth]{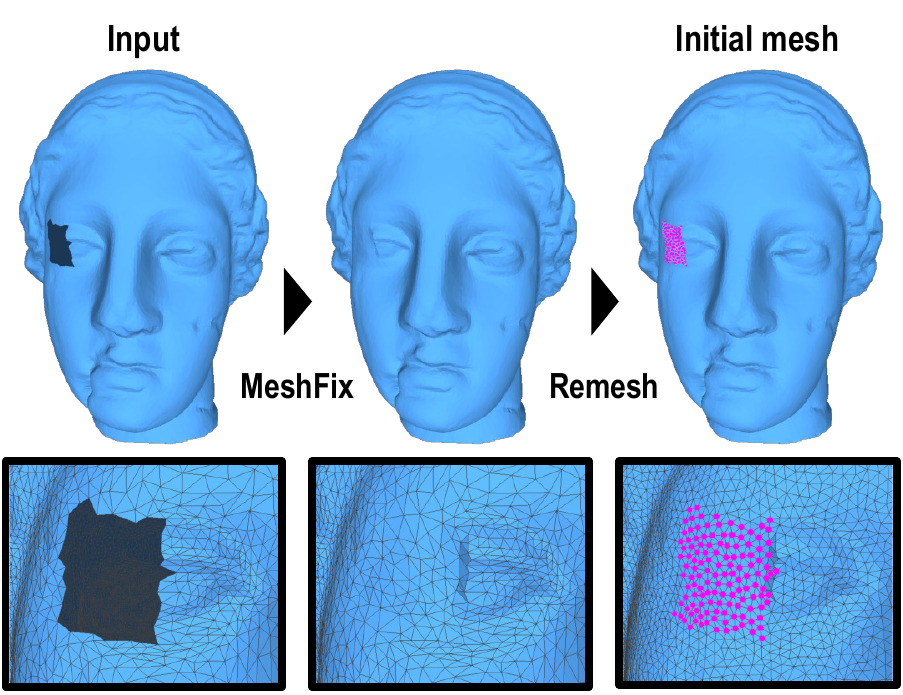}
  \caption{Generation of initial mesh $\mathcal{M}_{\text{init}}$. We first obtain a watertight manifold mesh using MeshFix~\cite{attene2010meshfix} (inserted vertices are colored pink) and then apply triangular remeshing~\cite{alliez2003remesh} to achieve uniform vertex density.}
  \label{fig:preprocess}
\end{figure}

The main difficulty of mesh inpainting is generating new vertices and faces in the missing regions. However, general GCNs, which we use in this study, have difficulty inserting new vertices and triangles and changing the vertex connectivity. Therefore, instead of inserting new vertices and faces using the neural network, we first convert $\mathcal{M}_{\text{in}}$ to a watertight manifold mesh using MeshFix~\cite{attene2010meshfix}, as shown in \cref{fig:preprocess}. Unfortunately, the densities between newly inserted vertices and originally existing vertices are inconsistent, which may worsen the performances of GCNs. To alleviate this problem, we obtain the initial mesh $\mathcal{M}_{\text{init}}$ with uniform vertex density by applying triangular remeshing~\cite{alliez2003remesh} after initial hole filling. After that, we train a neural network to learn the appropriate deformation of the initial mesh to obtain a complete mesh.

\begin{figure}[t!]
  \centering
  \includegraphics[width=\linewidth]{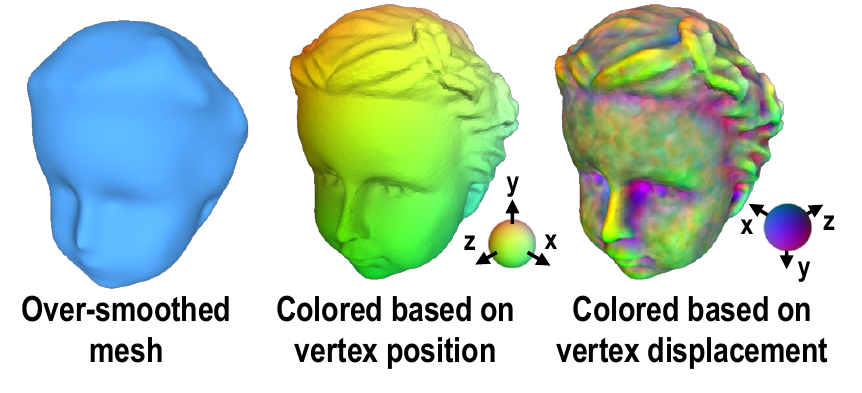}
  \caption{An oversmoothed mesh and meshes colored based on two geometric features. Local features are recognizable in the right mesh, colorized using vertex displacements, but not in the center, colored using vertex positions. This shows that local features can be extracted by subtracting vertex positions of the oversmoothed mesh from those of the original. The two color spheres in the center and right are identical but viewed from opposite sides to present the colors for outward and inward displacements.}
  \label{fig:pos-dis}
\end{figure}

In our inpainting, the GCN predicts the vertex displacements from the oversmoothed mesh rather than directly predicting the vertex positions. This is because a GCN can encode local geometric features more effectively when working with vertex displacements. This idea has recently been proposed by Hattori et al.~\cite{hattori2022ddmp} for mesh denoising, and they reported that it performed better when applying the DIP framework to mesh denoising. As shown in \cref{fig:pos-dis}, vertices with similar geometric features are colored similarly when the mesh is colored based on the vertex displacement. To extract these local geometric features, we apply 30-step uniform Laplacian smoothing without cotangent weights for the initial oversmoothing.

The above preprocessing, i.e., triangular remeshing and oversmoothing, alleviates the artifacts caused by initial hole filling and makes our method independent of the initial hole filling algorithm. Instead of other hole-filling methods (e.g., the advancing front method~\cite{zhao2007robust} used in a later approach~\cite{harary2014coherent} for the initial hole filling), we use MeshFix~\cite{attene2010meshfix} to ensure the manifoldness of the output hole-filled mesh.

\subsection{Network Architectures}
\label{ssec:network}

In this study, we consider two network architectures, SGCN and MGCN, to investigate the effect of network architecture on the mesh inpainting performance using a self-prior.

As its name indicates, SGCN performs the graph convolution defined on the mesh in the original resolution. SGCN consists of 13 graph convolution blocks and one vertex-wise fully connected layer. Each graph convolution block applies the Chebyshev spectral graph convolution (ChebConv)~\cite{defferrard2016chebnet} followed by batch normalization and a leaky rectified linear unit (LeakyReLU).

In contrast, MGCN employs an hourglass-type encoder--decoder network architecture. The encoder and decoder consist of several computation blocks. Each encoder block consists of five ChebConv layers with average mesh pooling, while each decoder block consists of five ChebConv layers with mesh unpooling. The last decoder block is followed by a single fully connected layer. The filter size of each ChebConv is determined by the order $K$ of the Chebyshev polynomial. We use $K=3$, corresponding to the convolution over vertices in 2-ring neighbors. Mesh pooling and unpooling are defined on a mesh using a precomputed static order of edge collapses, as described later. Due to space constraints, we provide additional details on the arrangement of these network components in the supplementary document.

Mesh pooling and unpooling required for MGCN have not been extensively explored, and to our knowledge, there is still no standard approach. While dynamic mesh pooling and unpooling of MeshCNN~\cite{hanocka2019meshcnn} are good choices when computation time is not a concern, we found its memory and time efficiency inadequate for processing large meshes with tens or hundreds of thousands of triangles. Instead, we define pooling and unpooling operations on a mesh using a precomputed series of edge collapses. For mesh pooling, we collapse edges in ascending order of quadratic error metrics (QEMs)~\cite{garland1997qem} until the number of vertices reaches the target. This process is similar to progressive meshes~\cite{hoppe1996progressive}. However, unlike the original approach, we add a penalty $P_{\text{val}}$ to the QEM of the $i$-th vertex to prevent its valence $n_{\text{val}}^i$ from being less than or equal to 3 after edge collapse, while also discouraging it from being more than 6.
\begin{equation}
  P_{\text{val}} = \begin{cases}
    \abs{n_{\text{val}}^i - 6} + 1 & (n_{\text{val}}^i > 3) \\
    \infty                         & (\text{otherwise})
  \end{cases}
\end{equation}
By introducing this penalty, we can obtain a simplified mesh with approximately regular triangulation without using either edge flip or edge split. In this way, our mesh pooling and unpooling can be defined simply by only considering pairs of vertices consecutively merged by edge collapses.

\begin{figure}[t!]
  \centering
  \includegraphics[width=\linewidth]{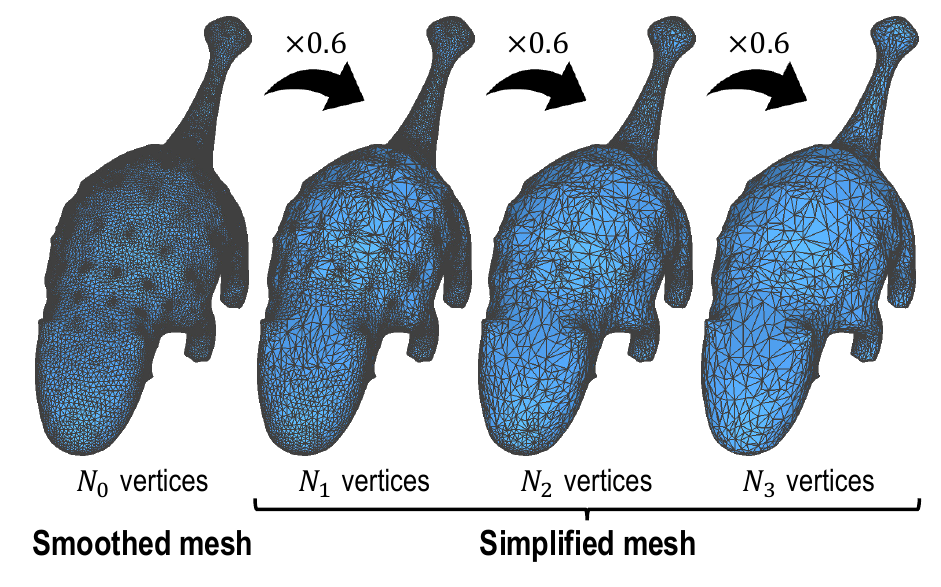}
  \caption{Meshes simplified using QEMs~\cite{garland1997qem} defined on a smoothed mesh. Each simplification reduces the number of vertices to $\SI{60}{\%}$, i.e., $N_{k} = \mathrm{round}(0.6 N_{k-1})$.}
  \label{fig:simplification}
\end{figure}

During mesh simplification, we record which vertex is merged with another vertex. When a set of vertices is merged to the $i$-th vertex, the average mesh pooling is defined simply as
\begin{equation}
  \vb{f}'_i = \frac{1}{\abs{\mathcal{C}(i)}} \sum_{k \in \mathcal{C}(i)} \vb{f}_k,
\end{equation}
where $\vb{f}$ and $\vb{f}'$ are the features before and after mesh pooling, respectively, and $\mathcal{C}(i)$ is an index set of the vertices merged to the $i$-th vertex. Conversely, mesh unpooling simply assigns the feature $\vb{f}_i$ back to the merged vertices with the indices in $\mathcal{C}(i)$. In our experiment, we precompute meshes at $R = 3$ different resolutions and define mesh pooling and unpooling between each pair of successive resolutions. To obtain a mesh with the next lower resolution, we reduce the number of vertices to $\SI{60}{\%}$. The simplified meshes with intermediate resolutions are shown in \cref{fig:simplification}.

\subsection{Self-Supervision}
\label{ssec:augmentation}

\begin{figure}[t!]
  \centering
  \includegraphics[width=\linewidth]{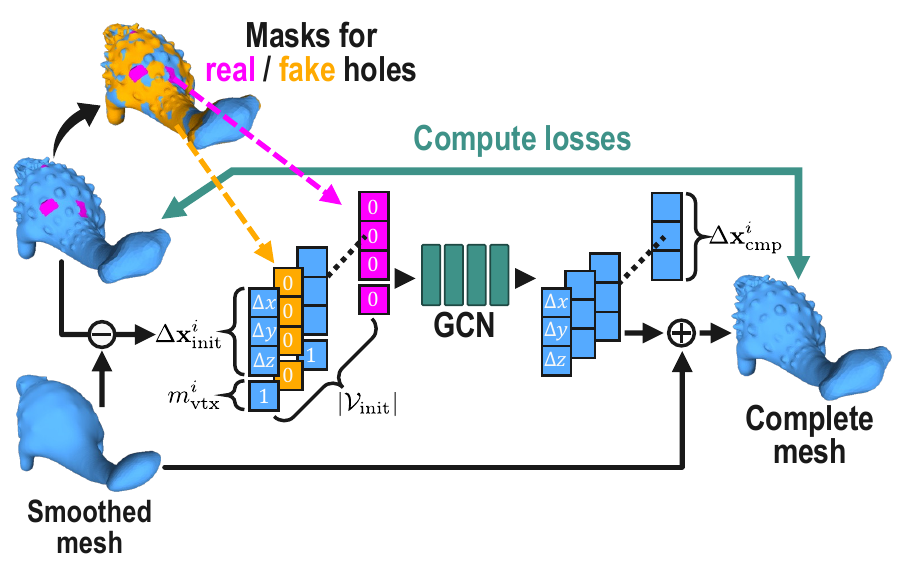}
  \caption{Self-supervision of the proposed GCN, where the GCN predicts the displacement of each vertex to reproduce the inpainted shape from the oversmoothed shape. We train the GCN in a self-supervised manner to predict the vertex displacements by filling zeros to the displacements at real and fake holes (colored pink and orange, respectively).}
  \label{fig:training}
\end{figure}

\Cref{fig:training} illustrates the proposed self-supervised learning for mesh inpainting. As shown in this figure, the GCN transforms a feature vector assigned to each vertex to another vector. At the $i$-th vertex, the GCN transforms a 4D vector $(\Delta x_{\text{in}}^i, \Delta y_{\text{in}}^i, \Delta z_{\text{in}}^i, m_{\text{vtx}}^i)$ to a 3D displacement vector $(\Delta x_{\text{out}}^i, \Delta y_{\text{out}}^i, \Delta z_{\text{out}}^i)$. The last entry $m_{\text{vtx}}^i \in \{ 0, 1 \}$ of the 4D input vector indicates the masking status of the $i$-th vertex, which takes 0 for vertices masked by either real or fake holes and takes 1 otherwise.

However, there are two problems in training the GCN to predict such displacements. First, the displacements at hole regions are unknown, and we cannot assess the accuracy of the displacements estimated by the GCN. Second, the first three entries of the input and output vectors (i.e., the input and output vertex displacements) at non-hole regions are always the same. This causes GCN training to get stuck in a trivial solution that merely represents an identity mapping.

To solve these problems, we propose an approach to training the GCN in a self-supervised manner by introducing fake holes as well as real holes. At the vertices covered by either fake or real holes, we mask the input displacements by zero vectors, i.e., $(\Delta x_{\text{in}}^i, \Delta y_{\text{in}}^i, \Delta z_{\text{in}}^i) = (0, 0, 0)$, and set the mask as $m_{\text{vtx}}^i = 0$. Then, the GCN is trained to predict the vertex displacements for the regions masked by real and fake holes using the known displacements in the non-hole regions. Then, the output displacements from the GCN are added to the oversmoothed mesh to obtain the output complete mesh $\mathcal{M}_{\text{cmp}}$. Because we know the correct displacement vectors at fake hole regions, we can penalize incorrect displacements using a loss function described later.

For self-supervision, we generate multiple sets of masks representing fake holes. For this purpose, we randomly select seed vertices with a certain probability $p$ and mark vertices inside the $k$-ring neighbor as included by fake holes. We set $p=\SI{1.4}{\%}$ and $k=4$ such that approximately $\SI{60}{\%}$ of the vertices are masked by fake holes. During this data augmentation, we select 40 sets of random seed vertices, thereby obtaining 40 meshes, each masked by a different set of fake holes.

\subsection{Loss Functions}
\label{ssec:losses}

To train both SGCN and MGCN, we use three kinds of loss terms: the data term for vertex positions $E_{\text{pos}}$, the data term for facet normals $E_{\text{nrm}}$, and the smoothness term for the directions of facet normals $E_{\text{reg}}$. In the following loss definitions, we simply denote the number of vertices and faces as $\abs{\mathcal{V}}$ and $\abs{\mathcal{F}}$, respectively, because $\abs{\mathcal{V}_{\text{init}}} = \abs{\mathcal{V}_{\text{cmp}}}$ and $\abs{\mathcal{F}_{\text{init}}} = \abs{\mathcal{F}_{\text{cmp}}}$.

\paragraph{Data term for vertex positions}

The error for $\vb{X}_{\text{cmp}}$, which is the output from the network, is calculated using the root mean squared error for the vertices that correspond to the original vertices of the initial hole-filled mesh $\mathcal{M}_{\text{init}}$. Therefore, we define the data term for vertex positions by introducing a mask $\tilde{m}_{\text{vtx}}^i \in \{ 0, 1 \}$, which takes 0 at vertices in real holes and 1 otherwise, to exclude the vertices in the \textit{real} hole regions:
\begin{equation}
  E_{\text{pos}}({\vb{X}}_{\text{cmp}},\!\vb{X}_{\text{init}}) \!=\! \qty( \frac{1}{\sum_i \tilde{m}_{\text{vtx}}^i } \sum_{i = 1}^{\abs{\mathcal{V}}} \tilde{m}_{\text{vtx}}^i \norm{{\vb{x}}_{\text{cmp}}^{i} \!-\! \vb{x}_{\text{init}}^{i}}_{2}^{2} )^{\frac{1}{2}},
  \label{eq:pos-loss}
\end{equation}
where $\norm{\,\cdot\,}_p$ is the $\ell_p$ norm of a vector. For multi-resolution meshes used by MGCN, the positional error $E_{\text{pos}}^{(r)}$ is equivalently defined for each resolution level $r=0, 1, \ldots$, and thus $E_{\text{pos}}$ = $E_{\text{pos}}^{(0)}$.

\paragraph{Data term for facet normals}

Facet normal directions are an important factor in accurately reconstructing local geometric features, such as sharp edges. Our network outputs vertex positions $\vb{X}_{\text{cmp}}$, and we can compute facet normals $\vb{N}_{\text{cmp}}$ using those vertex positions. To ensure that the facet normal directions of the output mesh match those of the triangles in the input mesh, we define the data term for facet normals by a mean absolute error for the triangles existing in $\mathcal{M}_{\text{in}}$ by introducing a mask $\tilde{m}_{\text{nrm}}^j \in \{0, 1\}$ that takes 0 at faces in \textit{real} holes and 1 otherwise.
\begin{equation}
  E_{\text{nrm}}({\vb{N}}_{\text{cmp}}, \vb{N}_{\text{init}}) = \frac{1}{\sum_j \tilde{m}_{\text{nrm}}^{j}} \sum_{j=1}^{\abs{\mathcal{F}}} \tilde{m}_{\text{nrm}}^{j} \norm{ \vb{n}_{\text{cmp}}^{j} - \vb{n}_{\text{init}}^{j}}_{1}.
  \label{eq:nrm-loss}
\end{equation}
Unlike $E_{\text{pos}}$, the data term $E_{\text{nrm}}$ for facet normals is computed for only the mesh at the highest resolution (i.e., $r = 0$), even for training MGCN, because of the different aims of these loss terms. The purpose of $E_{\text{nrm}}$ is to reproduce the sharp features of the input mesh, and thus, it does not work as we intend for meshes with lower resolutions where the sharp features of the original resolution may have vanished. Therefore, to save on computational cost, we compute $E_{\text{nrm}}$ for the mesh only with the finest resolution.

\paragraph{Smoothness term for facet normals}

As Mataev~et~al.~\cite{mataev2019deepred} reported, the regularization term defined with denoised output can prevent the DIP's neural network from overfitting to noise. In geometry processing, Hattori~et~al.~\cite{hattori2022ddmp} reported a similar finding that facet normal regularization using bilateral normal filtering (BNF)~\cite{zheng2010bilateral} is effective in preserving sharp geometric features such as edges and corners. Following those studies, we introduce a regularizer as the smoothness term for facet normals:
\begin{equation}
  E_{\text{reg}}(\vb{N}_{\text{cmp}}) = \frac{1}{|\mathcal{F}|} \sum_{j=1}^{|\mathcal{F}|} \norm{ \vb{n}_{\text{cmp}}^{j} - S_{\text{bnf}}^{(t)} (\vb{n}_{\text{cmp}}^{j}) }_{1},
  \label{eq:reg-loss}
\end{equation}
where $S_{\text{bnf}}^{(t)}$ denotes a smoothing function using BNF, and its superscript $t$ denotes that BNF is applied $t$ times. Our experiment shows that $t=5$ yields good results for CAD models. As with $E_{\text{nrm}}$, this smoothness term $E_{\text{reg}}$ for the facet normals is only computed for the mesh with the finest resolution for the same reason as explained in the previous paragraph. It is worth noting that $E_{\text{reg}}$ is computed only for a CAD model to encourage the recovery of sharp edges and corners (i.e., $\lambda_{\text{reg}} \neq 0$ only for CAD models, as described in the next paragraph).

\begin{table}[t!]
  \centering
  \caption{Weights for Loss Terms}
  \label{tab:weight-parameters}
  \begin{tblr}{
    colsep=2mm,
    colspec={Q[l]Q[l]X[c]X[c]X[c]X[c]X[c]X[c]}
    }
    \toprule
    {{{Network}}} & {{{Type}}} & {{{$\lambda_{\text{pos}}^{(0)}$}}} & {{{$\lambda_{\text{pos}}^{(1)}$}}} & {{{$\lambda_{\text{pos}}^{(2)}$}}} & {{{$\lambda_{\text{pos}}^{(3)}$}}} & {{{$\lambda_{\text{nrm}}$}}} & {{{$\lambda_{\text{reg}}$}}} \\
    \cmidrule[r]{1-1} \cmidrule[r]{2-2} \cmidrule[r]{3-6} \cmidrule{7-8}
    SGCN          & CAD        & 1.0                                & ---                                & ---                                & ---                                & 4.0                          & 4.0                          \\
                  & Non-CAD    & 1.0                                & ---                                & ---                                & ---                                & 1.0                          & 0.0                          \\
                  & Real scan  & 1.0                                & ---                                & ---                                & ---                                & 1.0                          & 0.0                          \\
    \cmidrule[r]{1-1} \cmidrule[r]{2-2} \cmidrule[r]{3-6} \cmidrule{7-8}
    MGCN          & CAD        & 0.35                               & 0.30                               & 0.20                               & 0.15                               & 4.0                          & 4.0                          \\
                  & Non-CAD    & 0.35                               & 0.30                               & 0.20                               & 0.15                               & 1.0                          & 0.0                          \\
                  & Real scan  & 0.35                               & 0.30                               & 0.20                               & 0.15                               & 1.0                          & 0.0                          \\
    \midrule
  \end{tblr}
\end{table}

\paragraph{Total loss}

Overall, we train the proposed networks using the following loss function:
\begin{equation}
  E = \sum_{r=0}^{R} \lambda_{\text{pos}}^{(r)} E_{\text{pos}}^{(r)} + \lambda_{\text{nrm}} E_{\text{nrm}} + \lambda_{\text{reg}} E_{\text{reg}},
  \label{eq:loss-function}
\end{equation}
where $R$ is the number of mesh resolutions, i.e., $R=1$ for SGCN and $R=3$ for MGCN, and each $\lambda$ modulates the influence of the respective loss term. The weights we used in the following experiments are summarized in \cref{tab:weight-parameters}. It is worth noting that the weights $\lambda_{\text{pos}}^{(r)}$ for training MGCN are set as $\sum_r \lambda_{\text{pos}}^{(r)}=1$ so that the positional error has an effect equivalent to that for SGCN.

\begin{figure}[t!]
  \centering
  \includegraphics[width=\linewidth]{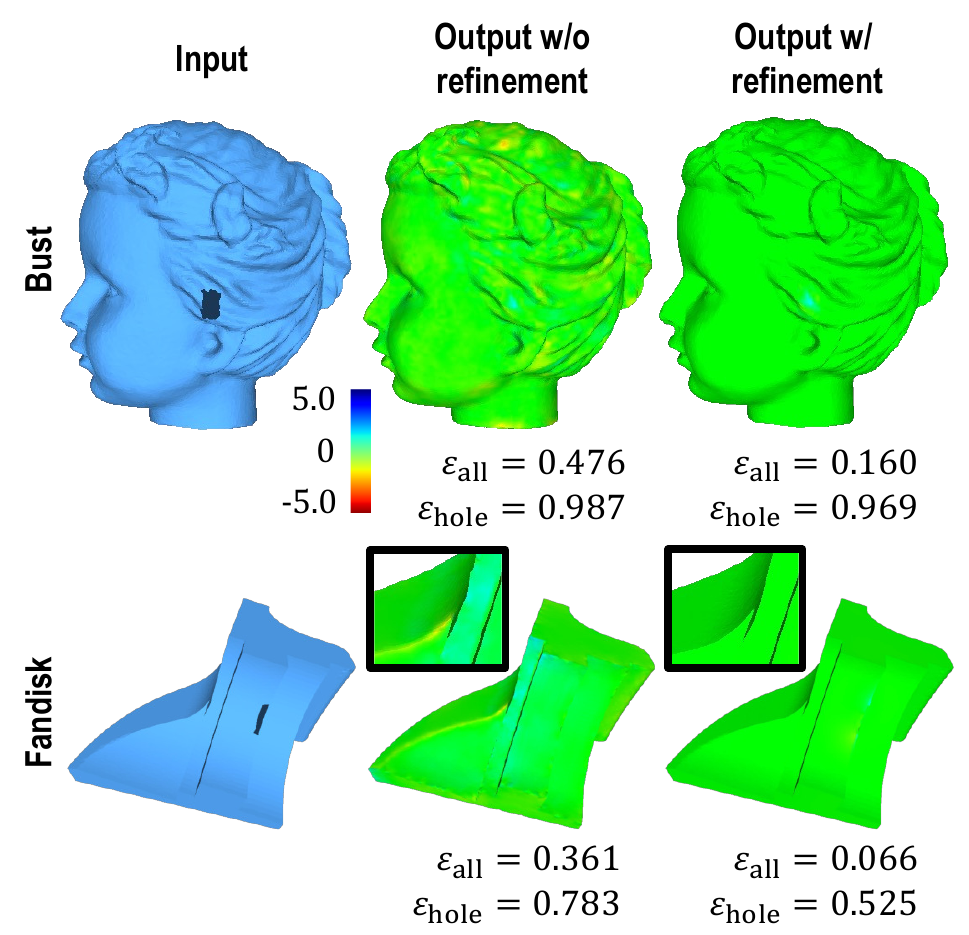}
  \caption{Effect of the refinement step for the outputs from MGCN. As this figure shows, the output shapes are significantly improved by the refinement step, as both $\epsilon_{\text{all}}$ and $\epsilon_{\text{hole}}$ (in the unit of $10^{-3}$) decrease significantly.}
  \label{fig:refinement}
\end{figure}

\subsection{Refinement Step}
\label{ssec:refinement}

As we mentioned previously, one of the problems of the DIP framework is the long processing time required to converge the optimization. We found that this problem is also common in polygonal meshes, and optimizing network parameters requires considerable computation time to obtain $\vb{X}_{\text{cmp}}$ that has vertex positions equivalent to those of $\vb{X}_{\text{init}}$ in non-hole regions. To alleviate this problem, we introduce a refinement step to restore the vertex positions in non-hole regions rather than waiting for the optimization to converge. Note that the refinement is only performed during the evaluation phase and not during the training phase.

For computational efficiency, we formulate this refinement step as a least-squares problem~\cite{lipman2004differential}. Let $\vb{L} = \vb{I}-\vb{D}^{-1}\vb{A} \in \mathbb{R}^{|\mathcal{V}|\times|\mathcal{V}|}$ be a Laplacian matrix of the mesh $\mathcal{M}_{\text{init}}$ where $\vb{I} \in \mathbb{R}^{|\mathcal{V}|\times|\mathcal{V}|}$ is an identity matrix, $\vb{A} \in \mathbb{R}^{|\mathcal{V}|\times|\mathcal{V}|}$ is an adjacency matrix, and $\vb{D} \in \mathbb{R}^{|\mathcal{V}|\times|\mathcal{V}|}$ is a diagonal degree matrix. Specifically, we solve the following minimization problem to obtain the final output.
\begin{equation}
  \vb{X}_{\text{out}} = \argmin_{\vb{X}} \frac{1}{2} \norm{ \vb{L} \qty( \vb{X} - \vb{X}_{\text{mix}} ) }_{F}^2 + \frac{\mu}{2} \norm{ \vb{Q} \odot \qty( \vb{X} - \vb{X}_{\text{init}} ) }_{F}^2.
  \label{eq:laplace-equation}
\end{equation}
Here, $\norm{\,\cdot\,}_{F}$ is the Frobenius norm of a matrix, and $\odot$ denotes a Hadamard product, i.e., an element-wise multiplication, of matrices. In this equation, $\vb{X}_{\text{mix}} \in \mathbb{R}^{|\mathcal{V}| \times 3}$ is a matrix where its $i$-th row $\vb{x}_{\text{mix}}^i$ is defined as
\begin{equation}
  \vb{x}_{\text{mix}}^i = \begin{cases}
    \vb{x}_{\text{init}}^i & (m_{\text{vtx}}^i = 1), \\
    \vb{x}_{\text{cmp}}^i  & (m_{\text{vtx}}^i = 0),
  \end{cases}
  \label{eq:mixed-vertices}
\end{equation}
and the masking matrix $\vb{Q} \in \mathbb{R}^{|\mathcal{V}| \times 3}$ masks the vertices at hole regions and their boundaries. The $i$-th row of $\vb{Q}$ is defined as
\begin{equation}
  \vb{q}^i = \begin{cases}
    (0, 0, 0) & (i \in \bar{\mathcal{H}}), \\
    (1, 1, 1) & (\text{otherwise}),
  \end{cases}
\end{equation}
where $\bar{\mathcal{H}}$ is a set of vertex indices that represent the union of vertices in hole regions $\mathcal{H}$ and its neighbors, i.e.,
\begin{equation}
  \bar{\mathcal{H}} = \qty{ i : \exists j \in \{ i \} \cup \mathcal{N}(i) \quad\text{s.t.}\quad m_{\text{vtx}}^j = 0 },
\end{equation}
where $\mathcal{N}(i)$ is an index set for vertices adjacent to the $i$-th vertex. We solve the minimization in \cref{eq:laplace-equation} as a sparse linear system, thereby obtaining the final refined output $\vb{X}_{\text{out}}$.

\Cref{fig:refinement} shows the effect of the refinement. In this figure, $\epsilon_{\text{hole}}$ represents the average positional error over only the hole regions, whereas $\epsilon_{\text{all}}$ represents that over the entire mesh. Their formal definitions are described later in \cref{sec:experiments}. As shown in the center of \cref{fig:refinement}, the complete mesh inferred by the proposed GCN entails small positional deviations at non-hole regions. In contrast, the refinement step successfully reduces the deviations at both the hole and non-hole regions as both $\epsilon_{\text{hole}}$ and $\epsilon_{\text{all}}$ decrease.

\begin{table*}[t!]
  \centering
  \caption{Quantitative Comparison of Inpainting Performance Using the Average Error Distance $\epsilon_{\text{hole}}$ in Units of $10^{-3}$}
  \label{tab:result-all}
  \begin{tblr}{
      width=\linewidth,
      colsep=3mm,
      colspec={
          X[l]
          Q[c]
          Q[c,si={table-format=2.3}]
          Q[c,si={table-format=2.3}]
          Q[c,si={table-format=2.3}]
          Q[c,si={table-format=2.3}]
          Q[c,si={table-format=2.3}]
          Q[c,si={table-format=2.3}]
          Q[c,si={table-format=1.3}]
        }
    }
    \toprule
                 & {{{Type}}} & {{{ADVF~\cite{zhao2007robust}}}} & {{{MFIX~\cite{attene2010meshfix}}}} & {{{SPSR~\cite{kazhdan2013screenedpoisson}}}} & {{{CCSC~\cite{harary2014coherent}}}} & {{{IPSR~\cite{hou2022iterative}}}} & {{{SGCN (ours)}}} & {{{{MGCN (ours)}}}} \\
    \cmidrule[r]{1-1} \cmidrule{2-2} \cmidrule[l]{3-9}
    CG           & CAD        & 13.900                           & 14.444                              & 11.141                                       & 10.641                               & 13.453                             & \BB 3.376         & 3.781               \\
    Fandisk      & CAD        & 5.692                            & 2.785                               & 3.346                                        & 4.944                                & 4.945                              & 1.890             & \BB 0.525           \\
    Part-lp      & CAD        & 9.464                            & 7.981                               & 6.524                                        & 8.476                                & 8.431                              & 1.543             & \BB 1.353           \\
    Sharp-sphere & CAD        & 5.322                            & 5.412                               & 6.432                                        & 5.762                                & 6.326                              & \BB 3.624         & 3.853               \\
    \cmidrule[r]{1-1} \cmidrule{2-2} \cmidrule[l]{3-9}
    Ankylosaurus & Non-CAD    & 2.282                            & 3.334                               & 1.364                                        & 1.869                                & 2.097                              & 1.871             & \BB 1.317           \\
    Bimba        & Non-CAD    & 3.470                            & 2.813                               & 2.780                                        & 4.428                                & 3.156                              & \BB 1.823         & 2.397               \\
    Bust         & Non-CAD    & 1.572                            & 1.277                               & 1.421                                        & 2.070                                & 1.642                              & \BB 0.942         & 0.969               \\
    Igea         & Non-CAD    & 2.458                            & 2.114                               & 1.607                                        & 1.411                                & 1.786                              & \BB 1.236         & 1.779               \\
    \cmidrule[r]{1-1} \cmidrule{2-2} \cmidrule[l]{3-9}
    Bunny        & Real scan  & 6.001                            & 5.551                               & 5.233                                        & 5.313                                & 5.862                              & \BB 5.224         & 5.299               \\
    Dragon       & Real scan  & 6.210                            & 3.232                               & 4.594                                        & 5.342                                & 7.502                              & \BB 2.269         & 2.565               \\
    \bottomrule
  \end{tblr}
\end{table*}

\section{Experiments}
\label{sec:experiments}

Our method is implemented using PyTorch and PyTorch Geometric. It was tested on a computer equipped with Intel Core i7 5930K CPU (\SI{3.5}{\GHz}, 6 cores), NVIDIA GeForce TITAN X GPU (\SI{12}{\GB} graphics memory), and \SI{64}{\GB} of RAM. We trained both SGCN and MGCN over 100 steps using the Adam optimizer with a learning rate of $\gamma=0.01$ and decay parameters $(\beta_1, \beta_2)=(0.9, 0.999)$, where the learning rate is halved after 50 steps.

For intuitive understanding, we colorize the output meshes based on the signed distance between the output and ground truth meshes. For quantitative evaluation, we also calculate the average Euclidean distances between the vertices of the output and ground truth meshes in two manners: one for all vertices and the other for those inserted into the missing regions. These average distances are denoted as $\epsilon_{\text{all}}$ and $\epsilon_{\text{hole}}$, respectively. The $\epsilon_{\text{all}}$ values are displayed only in \cref{fig:refinement} to show the effect of the refinement step, while the $\epsilon_{\text{hole}}$ values are shown in other figures and tables. These distances are normalized by the diagonal length of a bounding box of each mesh and are shown in units of $10^{-3}$.

We compared our results with those of previous methods, i.e., the advancing front method (ADVF)~\cite{zhao2007robust}, MeshFix (MFIX)~\cite{attene2010meshfix}, screened Poisson surface reconstruction (SPSR)~\cite{kazhdan2013screenedpoisson}, context-based coherent surface completion (CCSC)~\cite{harary2014coherent}, and iterative Poisson surface reconstruction (IPSR)~\cite{hou2022iterative}, all of which do not use datasets and work only with an input mesh with holes. In this comparison, we used 13 meshes, including CAD, non-CAD, and real-scanned models. The quantitative evaluations were conducted only for 10 meshes with known ground truth geometries and are summarized in \cref{tab:result-all}.

\begin{figure}[t!]
  \centering
  \includegraphics[width=\linewidth]{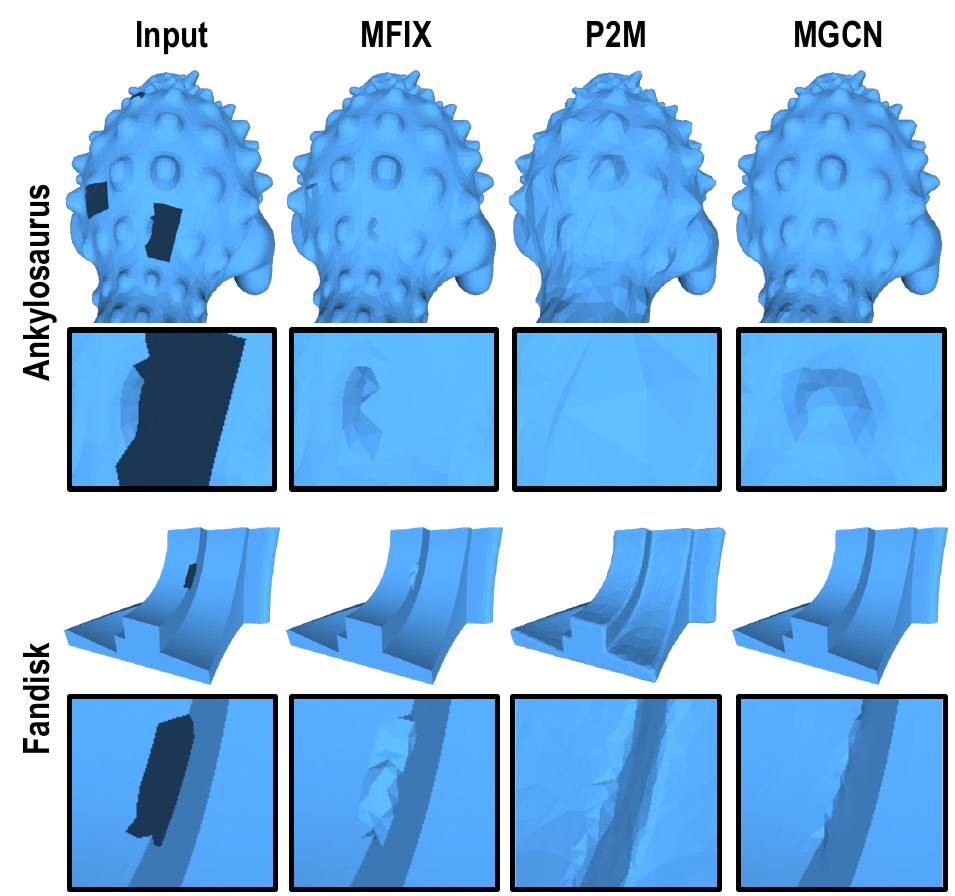}
  \caption{Inpainting results for selected models, obtained by MeshFix (our baseline), Point2Mesh~\cite{hanocka2020point2mesh}, and our method.}
  \label{fig:result-p2m}
\end{figure}

\begin{figure*}[t!]
  \centering
  \includegraphics[width=\linewidth]{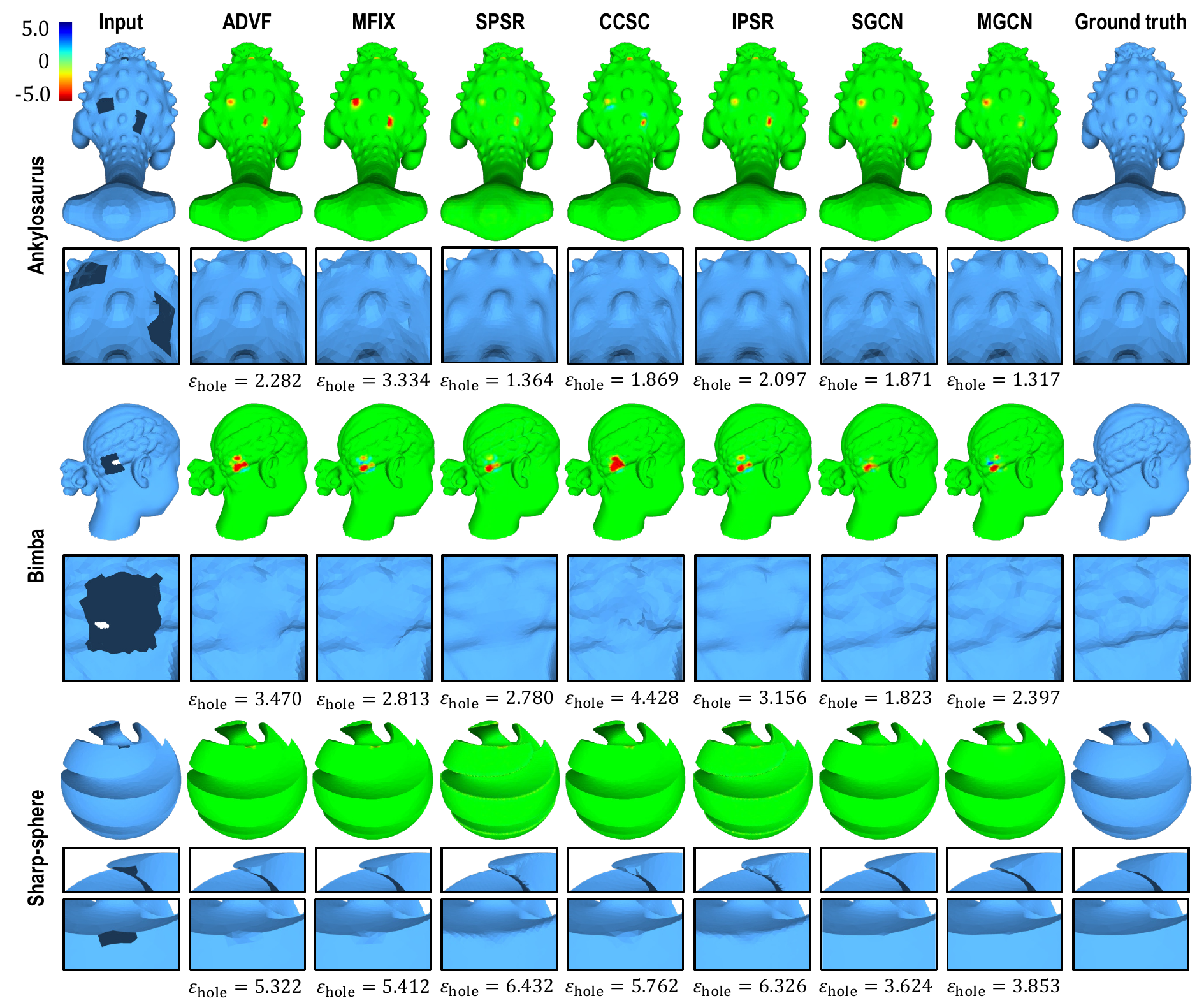}
  \captionof{figure}{Inpainting performances for meshes with artificial holes. The output meshes are colored based on the distance to the ground truth, and the average distance from the output mesh at holes $\epsilon_{\text{hole}}$ is displayed below each mesh. Note that the positional error for MeshFix is computed after the hole-filled regions are remeshed.}
  \label{fig:result-artificial}
\end{figure*}

In addition to the traditional approaches above, we compared the result from MGCN and that from Point2Mesh (P2M)~\cite{hanocka2020point2mesh}, a self-supervised surface reconstruction method similar to ours. As \cref{fig:result-p2m} shows, P2M appears capable of capturing the self-similarity of the input geometry. However, in our testing environment, P2M was only applicable to low-resolution meshes because MeshCNN~\cite{hanocka2019meshcnn}, its backend network, requires a large amount of graphics memory. In contrast, our method employs memory-efficient GCNs as a backend and has successfully restored characteristic shape features (e.g., repetitive bumps on the back of the ankylosaurus) and sharp edges (e.g., those of fandisk) while maintaining mesh resolution.

Although we also compared our method with recent data-driven shape completion methods (i.e., PMP-Net~\cite{wen2021pmp}, IF-Net~\cite{chibane2020implicit}, and ShapeFormer~\cite{yan2022shapeformer}), we found that the results of these methods are strongly affected by shapes in the training datasets. Therefore, in the following, we only show the comparison of the proposed methods with traditional methods that do not use datasets. Refer to the supplementary document for the comparison with these data-driven methods.

\subsection{Results on CAD and Non-CAD Models}

The top eight rows of \cref{tab:result-all} show the results of quantitative analysis on the CAD and non-CAD models. As these models do not originally have holes, we made several ``artificial'' holes and considered them to be real holes. The results for some of these models are shown in \cref{fig:result-artificial}. As shown in this figure, traditional methods, such as ADVF~\cite{zhao2007robust} and MFIX~\cite{attene2010meshfix}, are prone to inpaint holes with a planar shape because they are based solely on local geometry processing. SPSR~\cite{kazhdan2013screenedpoisson} and IPSR~\cite{hou2022iterative}, the variants of Poisson surface reconstruction~\cite{kazhdan2006poisson}, often inpaint holes with oversmoothed patches and fail to recover sharp features, such as edges and corners. CCSC~\cite{harary2014coherent} appears to be able to recover the characteristic object shapes. However, we tested CCSC with several meshes and found that its results did not always match the shapes around holes. Additionally, the results of the quantitative evaluation for CCSC were not as good as that of other methods (see \cref{tab:result-all}), although CCSC was able to insert more complicated patches compared to the previous methods. In contrast to these previous methods, our method successfully restores missing details and sharp features, and either SGCN or MGCN achieved the lowest $\epsilon_{\text{hole}}$ for all the meshes.

\begin{figure*}[t!]
  \centering
  \includegraphics[width=\linewidth]{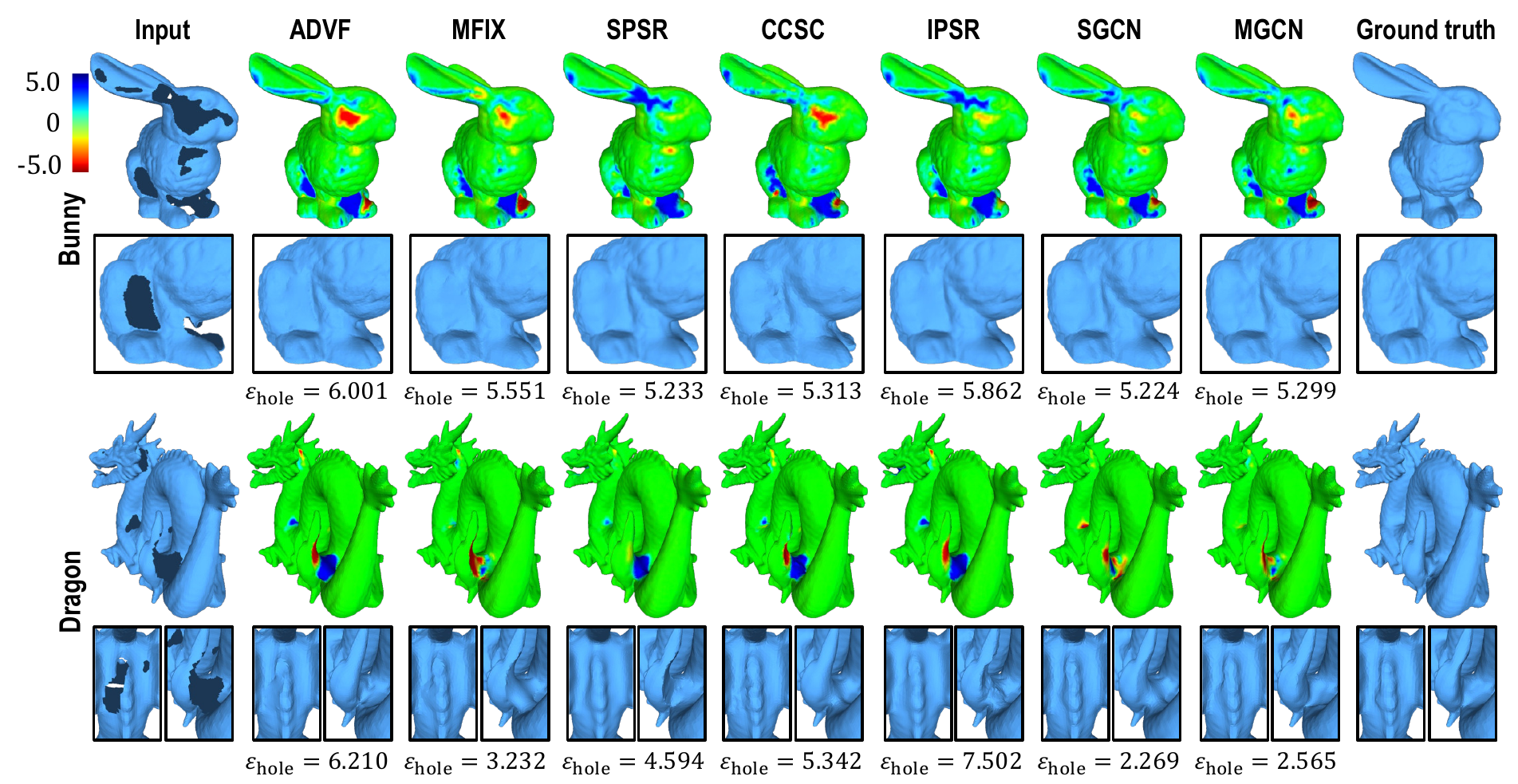}
  \caption{Inpainting performances for real scans provided by the Stanford 3D Repository~\cite{stanford3d}.}
  \label{fig:result-real}
\end{figure*}

\begin{figure}[t!]
  \centering
  \includegraphics[width=\linewidth]{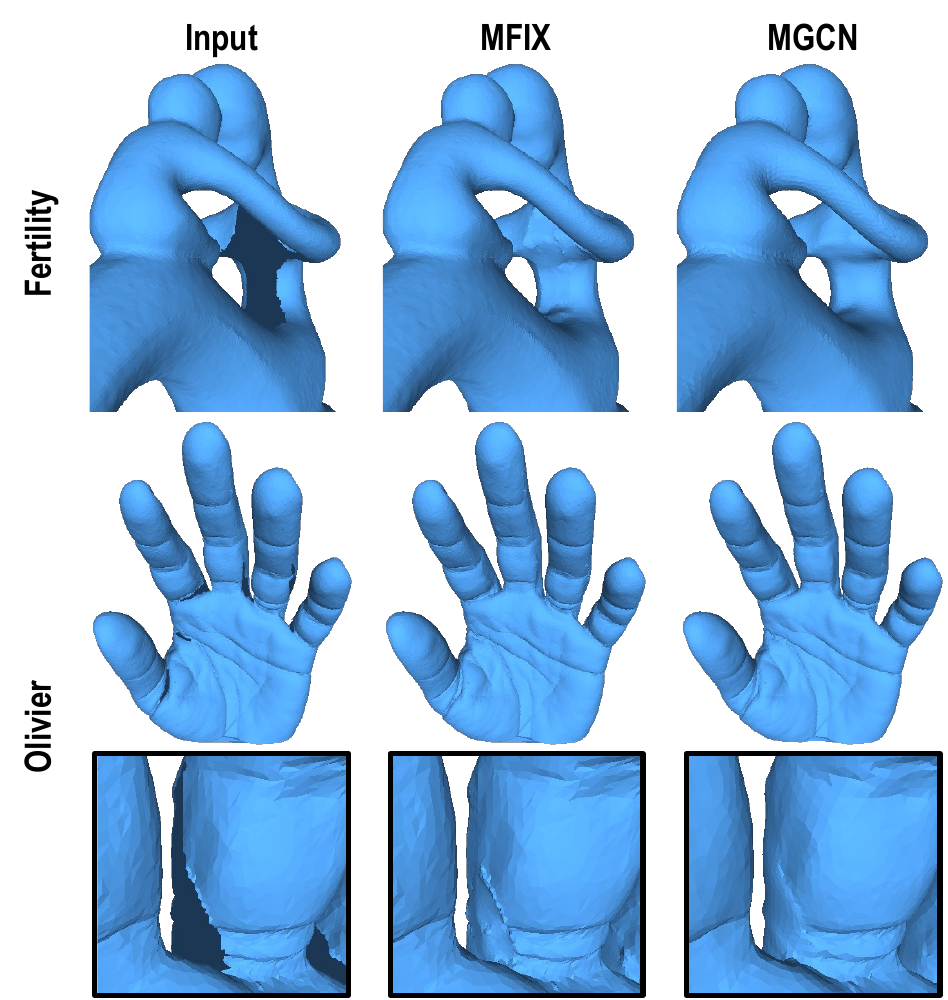}
  \caption{Inpainting results for real scans distributed by the AIM@SHAPE Shape Repository~\cite{aim3d}.}
  \label{fig:result-aim}
\end{figure}

\begin{figure}[t!]
  \centering
  \includegraphics[width=\linewidth]{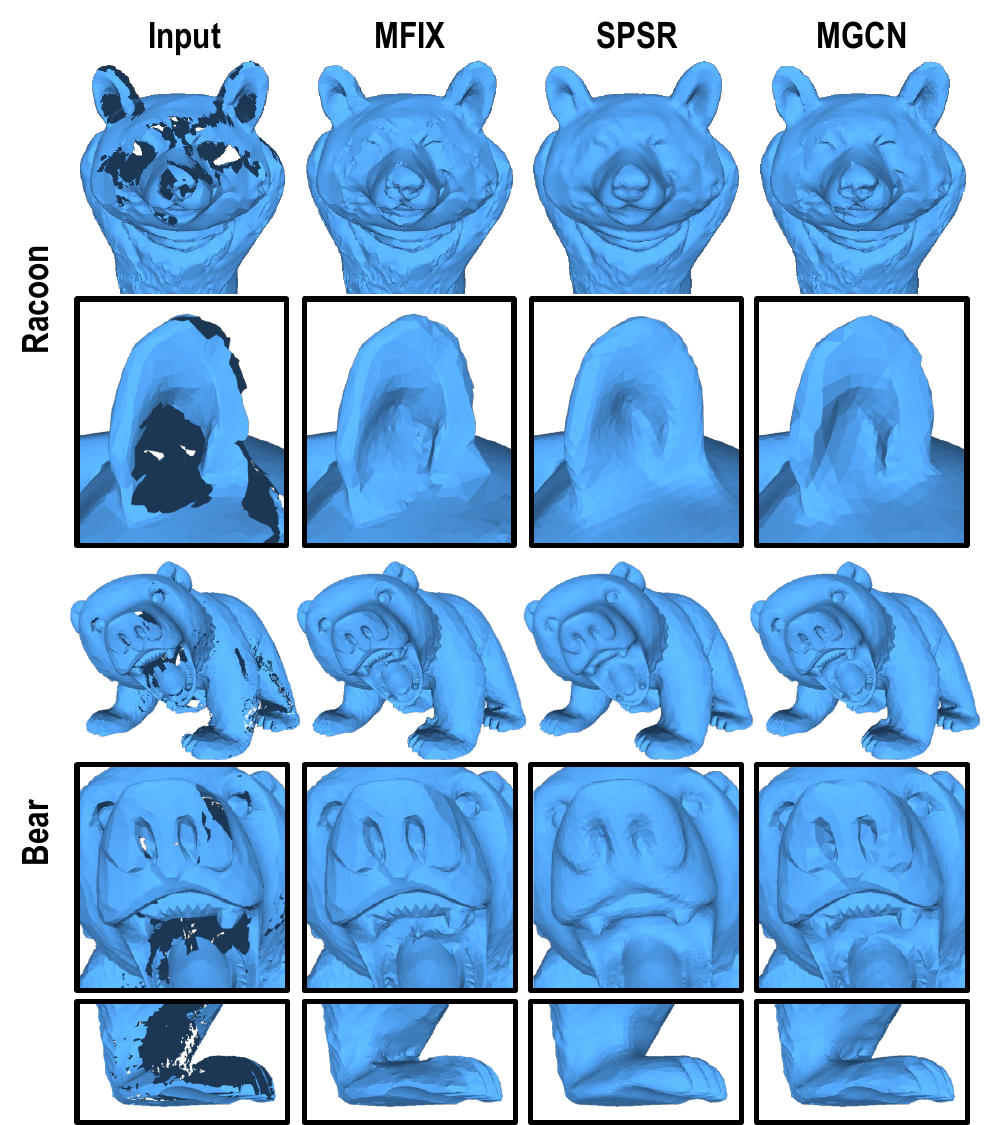}
  \caption{Inpainting results for real-scanned meshes obtained using an optical 3D scanner (GOM ATOS Core 135).}
  \label{fig:result-atos}
\end{figure}

\subsection{Results on Real-Scanned Models}

We evaluated our method using real-scanned models provided by the Stanford 3D Repository~\cite{stanford3d}, where incomplete range scans are provided for each model. Using a set of range scans of each model, we synthesized an input mesh with missing regions by excluding some of the range scans and synthesized its corresponding complete mesh using all the range scans. Specifically, we excluded 6/10 and 15/45 range scans for bunny and dragon models, respectively. Since the range scans are provided as point clouds, we obtained the surface geometries of incomplete and complete meshes using SPSR~\cite{kazhdan2013screenedpoisson}. Although SPSR may fill the missing regions of an incomplete point cloud, we identified the missing regions by thresholding the triangle sizes.

\Cref{fig:result-real} shows the inpainting results for the models from the Stanford 3D Repository, which shows that these real-scanned models contain large holes with complicated shapes. In particular, a large hole between the bunny's front legs cannot be inpainted by any of the methods, while SGCN achieves among the lowest $\epsilon_{\text{hole}}$. In contrast, for the dragon model, both our SGCN and MGCN achieve significantly lower $\epsilon_{\text{hole}}$ than all the other methods, naturally reproducing the characteristic shapes of the crest on the dragon's back and the scaly skin texture on its body.

We also examined the performance of our method with a more practical scenario and compared the results with MFIX~\cite{attene2010meshfix} as a baseline, which we used for the initial hole filling. \Cref{fig:result-aim} shows the results for the real scans provided by the AIM@SHAPE Shape Repository~\cite{aim3d}. For the fertility model, we generated the incomplete input mesh as performed for the models in the Stanford 3D Repository. For the Olivier model, we directly applied our method to the provided mesh that originally had missing regions at occluded parts between fingers. As \cref{fig:result-aim} shows, MFIX fills the holes well in that the plausible shapes are reproduced, while our method appropriately moves the vertices to reproduce the object shapes more naturally.

\Cref{fig:result-atos} shows the results for two real-scanned models, i.e., a plastic raccoon toy and a wooden bear carving, which we obtained using an optical 3D scanner (GOM ATOS Core 135) by ourselves. The raccoon sample has many complicated holes because of self-occlusion at the ears and the black color around the eyes. Similarly, the bear sample has complicated holes, particularly inside the mouth and knees of the rear feet, due to self-occlusion. Even for these challenging samples, the results of our method are more visually adequate than those of baseline methods. MFIX could fill the holes, but the geometries of the filled parts appear somewhat unnatural, as we can observe in enlarged images of the ear and the regions inside the bear's mouth in \cref{fig:result-atos}. Furthermore, our method successfully reproduced finer details compared with SPSR. While SPSR could also fill the holes and its results appear visually adequate, details like the bear's teeth have been lost. As evident from these results, our self-prior-based method achieves both visual adequateness and accurate reproduction of fine details.

\begin{figure*}[t!]
  \centering
  \includegraphics[width=\linewidth]{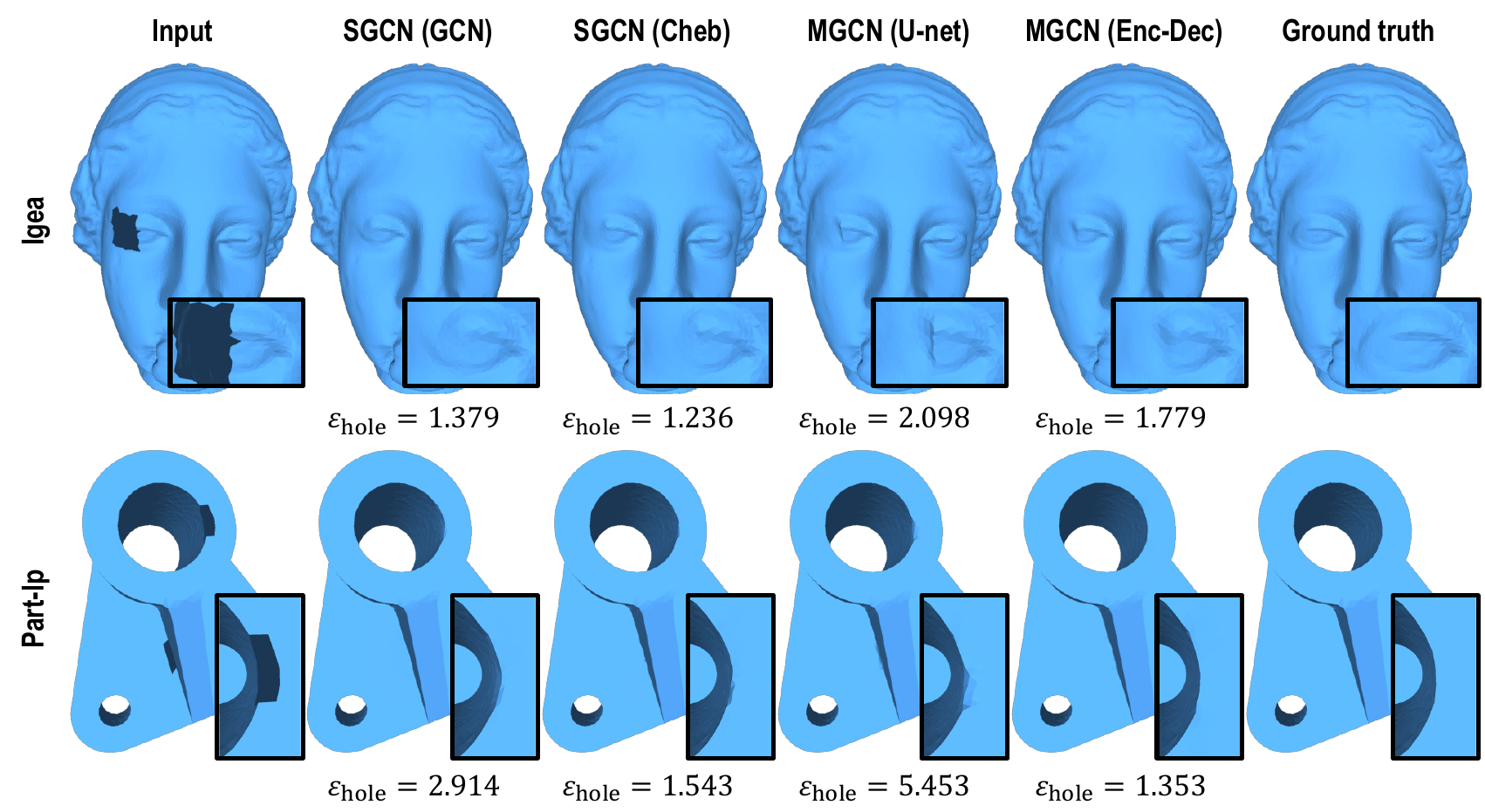}
  \caption{Inpainted meshes obtained with different network architectures.}
  \label{fig:result-network}
\end{figure*}

\begin{table}[t!]
  \centering
  \caption{Inpainting Performances of MeshFix and Our Methods Using Different Network Architectures}
  \label{tab:result-network}
  \begin{tblr}{
      colsep=2mm,
      colspec={
          X[l]
          Q[c,si={table-format=1.3}]
          Q[c,si={table-format=1.3}]
          Q[c,si={table-format=1.3}]
          Q[c,si={table-format=1.3}]
          Q[c,si={table-format=1.3}]
        }
    }
    \toprule
                                                                                                     & {{{MFIX}}} & {{{SGCN                                 \\(GCN)}}} & {{{SGCN\\(Cheb)}}} & {{{MGCN\\(U-net)}}} & {{{MGCN\\(Enc--Dec)}}} \\
    \cmidrule{1-1} \cmidrule[l]{2-6}
    Ankylo.                                                                                          & 3.334      & 1.875   & 1.871     & 2.492 & \BB 1.317 \\
    Igea                                                                                             & 2.114      & 1.379   & \BB 1.236 & 2.098 & 1.779     \\
    Fandisk                                                                                          & 2.785      & 1.377   & 1.890     & 3.527 & \BB 0.525 \\
    Part-lp                                                                                          & 7.981      & 2.914   & 1.543     & 5.453 & \BB 1.353 \\
    \bottomrule
    \SetCell[c=6]{c} $^*$The values in this table are $\epsilon_{\text{hole}}$ in units of $10^{-3}$ &            &         &           &       &           \\
  \end{tblr}
\end{table}

\begin{figure}[t!]
  \centering
  \includegraphics[width=\linewidth]{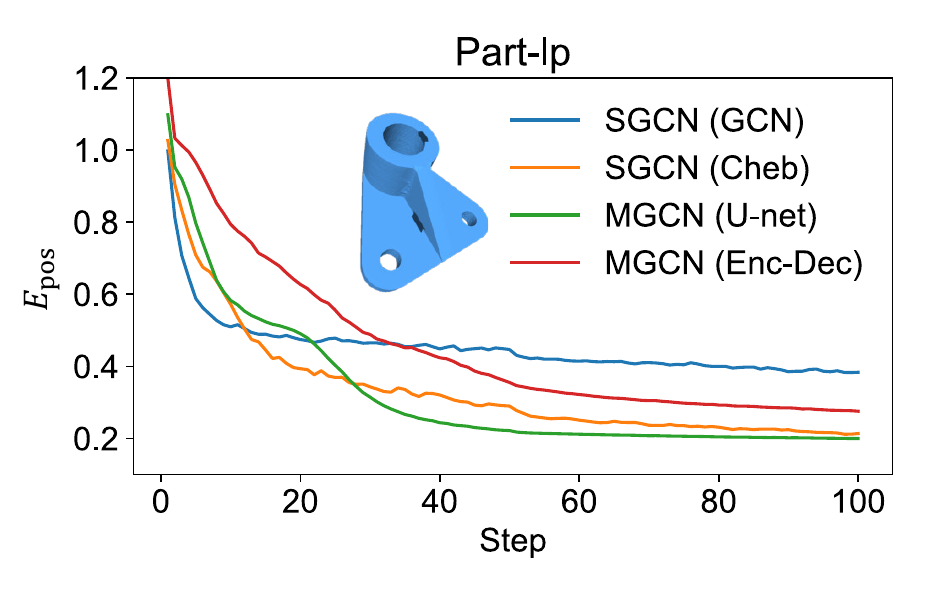}
  \caption{Speed of $E_{\text{pos}}$ convergence for our methods using different network architectures.}
  \label{fig:loss-convergence}
\end{figure}

\subsection{Comparison of Different Network Architectures}

We now investigate other possible structures of SGCN and MGCN. While we constructed both SGCN and MGCN using ChebConv~\cite{defferrard2016chebnet} as a graph convolution layer, we may also be able to employ GCNConv, the more efficient graph convolution, proposed by Kipf and Welling~\cite{kipf2017gcnconv}, which is widely known as a standard graph convolution layer. GCNConv is regarded as an approximation of ChebConv with a Chebyshev polynomial of order $K=2$ and is expected to perform similarly. Furthermore, since MGCN employs an hourglass-type encoder--decoder structure, we may also introduce skip connections to make it resemble U-net~\cite{ronneberger2015unet} and multi-resolution CNN~\cite{wang2020multi}, which have proven effective in deep feature extraction for various applications.

\Cref{fig:result-network,tab:result-network} show the visual appearances and quantitative evaluations of the results obtained by different network architectures. In comparing the two types of SGCN, one with GCNConv (i.e., SGCN (GCN)) and the other with ChebConv (i.e., SGCN (Cheb)), their performances are found to be equivalent. However, we observed an interesting behavior: the training process, specifically the network parameter optimization, was more stable when using ChebConv rather than GCNConv. This behavior is evident in \cref{tab:result-network}, where $\epsilon_{\text{hole}}$ values are consistently low for various inputs. Moreover, the instability problem of GCNConv becomes more serious when used with MGCN. In particular, the training process of MGCN with GCNConv did not converge to any reasonable solutions. Because of this, the results of ``MGCN (GCN)'' are excluded from \cref{fig:result-network,tab:result-network}.

The effect of skip connections on MGCN can be observed in \cref{tab:result-network}, which shows that ``MGCN (Enc--Dec),'' a simple encoder--decoder architecture without skip connections, outperforms ``MGCN (U-net)'' with skip connections, similar to the U-net architecture. Such a behavior is also reported in the original DIP paper~\cite{ulyanov2018dip}. Our result suggests that the unwanted effect of skip connections is also apparent when applying the DIP framework to polygonal meshes.

When it comes to the relationship between convergence speed and output quality, a network architecture exhibiting faster training convergence is found to obtain better results. For example, \cref{fig:loss-convergence} shows that training of ``SGCN~(Cheb)'' converges faster than that of ``SGCN~(GCN),'' and the former obtains better results, as described above. However, as we discussed previously, skip connections do not improve the inpainting performance despite the faster convergence of ``MGCN~(U-net)'' than ``MGCN~(Enc--Dec).'' During our experiment, we observed that networks with skip connections are prone to becoming stuck in a trivial solution representing an identity mapping. Based on these observations, we speculate that this issue may arise from the difference in inductive bias introduced by the networks with and without skip connections.

\begin{table}[t!]
  \centering
  \caption{Running Times for Loss Convergence by Different Network Architectures}
  \label{tab:timing}
  \begin{tblr}{
    width=\linewidth,
    colspec={Q[l]Q[c]Q[c]X[c]X[c]X[c]},
    colsep=2mm,
    }
    \toprule
            &                &             & \SetCell[c=3]{c} {{{Time}}}                          &                                                                                                 \\
    \cmidrule{4-6}
            &                &             & \SetCell[r=3]{c} {{{SGCN (GCN) \textbf{300} steps}}} & \SetCell[r=3]{c} {{{SGCN (Cheb) 100 steps}}} & \SetCell[r=3]{c} {{{MGCN (Enc--Dec) 100 steps}}} \\
            &                &             &                                                      &                                              &                                                  \\
            & {{{Vertices}}} & {{{Faces}}} &                                                      &                                              &                                                  \\
    \cmidrule[r]{1-1} \cmidrule[r]{2-3} \cmidrule{4-7}
    Ankylo. & 21090          & 42176       & \sitime{42}{21}                                      & \sitime{28}{43}                              & \sitime{24}{21}                                  \\
    Igea    & 44143          & 88282       & \sitime{93}{16}                                      & \sitime{61}{56}                              & \sitime{48}{56}                                  \\ Fandisk & 16724 & 33444 & \sitime{37}{47} & \sitime{24}{17} & \sitime{19}{42} \\ Part-lp & 10165 & 20338 & \sitime{22}{09} & \sitime{14}{13} & \sitime{13}{29} \\ \bottomrule
  \end{tblr}
\end{table}

\Cref{tab:timing} shows the computation times for different network architectures needed to iterate the optimization over 100 steps; ``SGCN~(Cheb)'' and ``MGCN~(Enc--Dec)'' exhibit almost equal time complexity. As shown in \cref{fig:loss-convergence}, the network with GCNConv converges more slowly, but the time required for forward and backward network evaluations is shorter than that with ChebConv. However, when SGCN utilizes GCNConv, it requires three times the number of steps (i.e., 300 steps) to achieve a loss value as small as the one attained by SGCN employing ChebConv. As a result, the total time required to achieve the same loss value is longer with GCNConv, as shown in \cref{tab:timing}. Thus, ChebConv is a preferable choice for graph convolution in terms of time performance as well. It should be noted that incorporating skip connections into MGCN had a negligible effect on the computation time.

Currently, the time required by our method remains longer than that required by traditional approaches. This is a typical problem of self-prior-based methods, reflecting the fact that these techniques are still evolving. For instance, even CCSC~\cite{harary2014coherent}, one of the traditional methods requiring comparatively long processing time, takes no more than ten minutes for an input mesh with several tens of thousands of triangles (e.g., it took about 7 minutes for the Igea model with approximately 90000 triangles). Although our method is still slower than even CCSC, it exhibits a significant speed advantage over P2M~\cite{hanocka2020point2mesh}, a recent self-prior-based approach. In our testing environment, P2M required about two hours to process an input mesh with approximately ten thousand triangles.

\subsection{Effect of Error Terms}

\begin{table}[t!]
  \centering
  \caption{Results of Ablation Study for MGCN Losses}
  \label{tab:result-loss}
  \begin{tblr}{
    width=\linewidth,
    colsep=1.5mm,
    colspec={Q[l]X[c]X[c]Q[c]Q[c]Q[c]},
    }
    \toprule
                               & {{{MFIX}}} & \BB {{{MGCN}}} & {{{MGCN                              \\w/o $E_{\text{pos}}^{(1,2,3)}$}}} & {{{MGCN\\w/o $E_{\text{reg}}$}}} & {{{MGCN\\w/o $E_{\text{nrm}}$}}} \\
    \cmidrule{1-1} \cmidrule[l]{2-2} \cmidrule[l]{3-6}
    $E_{\text{pos}}^{(0)}$     & ---        & \checkmark     & \checkmark & \checkmark & \checkmark \\
    $E_{\text{pos}}^{(1,2,3)}$ & ---        & \checkmark     & $\times$   & \checkmark & \checkmark \\
    $E_{\text{nrm}}$           & ---        & \checkmark     & \checkmark & \checkmark & $\times$   \\
    $E_{\text{reg}}$           & ---        & \checkmark     & \checkmark & $\times$   & $\times$   \\
    \cmidrule{1-1} \cmidrule[l]{2-2} \cmidrule[l]{3-6}
    Ankylo.                    & 3.334      & \BB 1.317      & 1.533      & ---        & 1.619      \\
    Igea                       & 2.114      & \BB 1.779      & 1.949      & ---        & 1.952      \\ Fandisk & 2.785 & \BB 0.525 & 0.717 & 0.568 & 2.289 \\ Part-lp & 8.035 & \BB 1.353 & 1.501 & 2.165 & 7.299 \\ \bottomrule \SetCell[c=6]{c} $^*$The values in this table are $\epsilon_{\text{hole}}$ in units of $10^{-3}$ & & & & & \\
  \end{tblr}
\end{table}

\begin{figure}[t!]
  \centering
  \includegraphics[width=\linewidth]{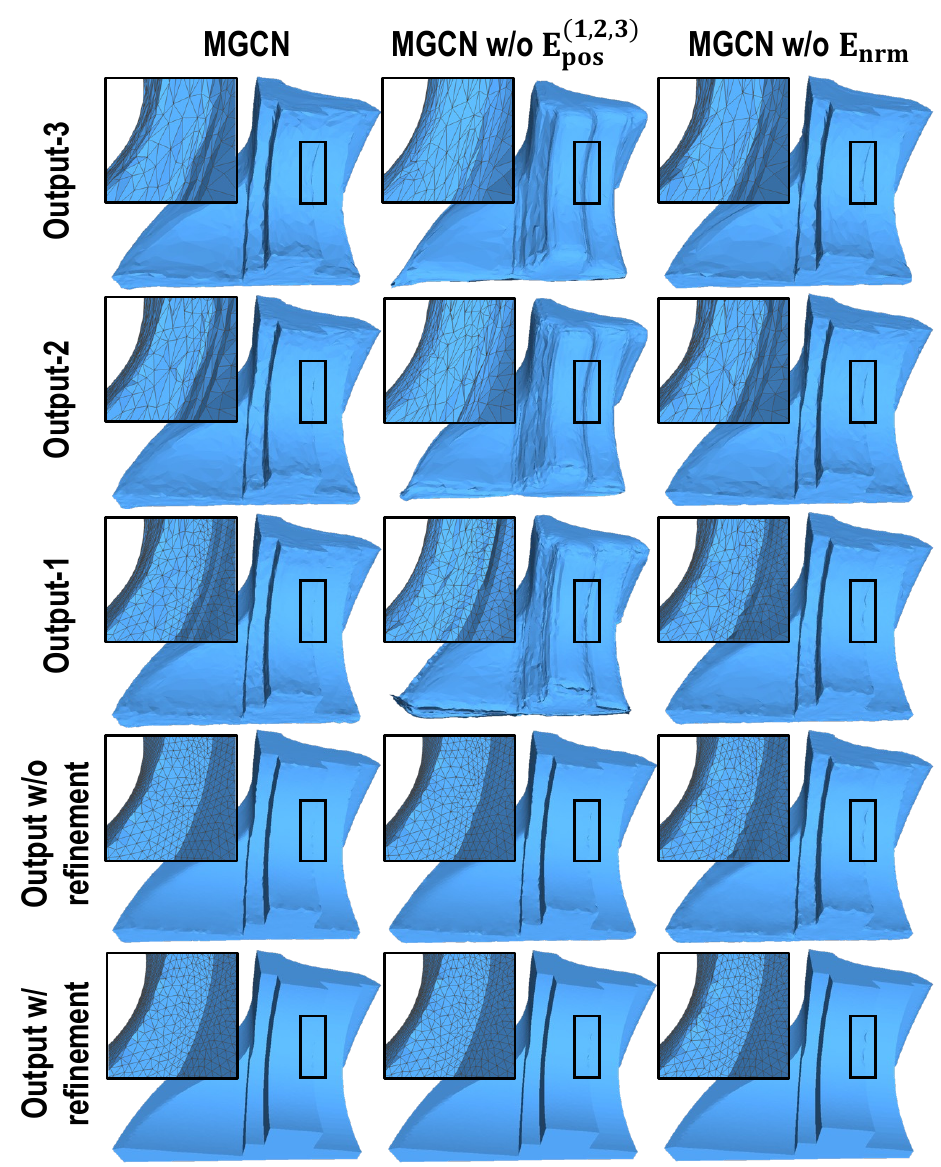}
  \caption{Inpainted meshes at different resolutions. While the outputs after the refinement step appear nearly identical, there are subtle differences in the shapes of areas that were initially holes (best viewed on screen).}
  \label{fig:result-ablation}
\end{figure}

\Cref{tab:result-loss} shows the $\epsilon_{\text{hole}}$ values obtained by an ablation study on error terms used to train MGCN. In this table, ablated losses for each MGCN variant are marked by a cross mark ``$\times$'', while remaining losses are marked by a check mark ``\checkmark.'' It should be noted that the label, ``w/o $E_{\text{nrm}}$,'' denotes that the corresponding MGCN variant does not consider the deviations of normal vectors. In other words, it neither employs $E_{\text{nrm}}$ nor $E_{\text{reg}}$, as indicated in \cref{tab:result-loss}. In addition to this table, \cref{fig:result-ablation} shows the results of visual analyses on the output meshes with different resolutions. In \cref{tab:result-loss,fig:result-ablation}, we conducted each analysis by excluding one of the following error terms: the sum of a part of multi-resolution reconstruction terms $E_{\text{pos}}^{(1,2,3)} \triangleq \sum_{r=1}^{3} E_{\text{pos}}^{(r)}$, the normal reconstruction term $E_{\text{nrm}}$, or the regularization term $E_{\text{reg}}$. Since $E_{\text{reg}}$ is not used for non-CAD models, i.e., ankylosaurus and Igea models, horizontal lines are drawn in the respective cells.

As demonstrated visually in \cref{fig:result-ablation}, excluding $E_{\text{pos}}^{(1,2,3)}$ leads the network to neglect the quality of intermediate low-resolution meshes. As the values in \cref{tab:result-loss} indicate, this exclusion results in a compromised reproduction of the vertex positions at the original resolution. Moreover, when $E_{\text{nrm}}$ is excluded, the reproduction of sharp features, such as edges and corners, at hole regions is limited, although those at non-hole regions can be reproduced properly by the refinement step, as shown in \cref{fig:result-ablation}. The result also worsens quantitatively, as shown in \cref{tab:result-loss}. In addition, \cref{tab:result-loss} shows that $\epsilon_{\text{hole}}$ deteriorates by ablating the regularizer, which suggests that the facet-normal regularizer $E_{\text{reg}}$ is important for reproducing the sharp features. Thus, all the error terms used to train MGCN are essential to maximizing the performance of the proposed method.

\begin{table}[t!]
  \centering
  \caption{Effect of Fake Hole Size}
  \begin{tblr}{
    width=0.85\linewidth,
    colsep=2mm,
    colspec={Q[l]X[c]X[c]X[c]}
    }
    \toprule
                                                                                                     & {{{$k=2$}}} & {{{$k=4$}}} & {{{$k=6$}}} \\
    \cmidrule{1-1} \cmidrule[l]{2-4}
    Ankylosaurus                                                                                     & 1.689       & \BB 1.317   & 1.393       \\
    Igea                                                                                             & 1.946       & 1.779       & \BB 1.677   \\
    Fandisk                                                                                          & \BB 0.487   & 0.525       & 0.719       \\
    Part-lp                                                                                          & \BB 1.220   & 1.353       & 2.685       \\
    \bottomrule
    \SetCell[c=4]{c} $^*$The values in this table are $\epsilon_{\text{hole}}$ in units of $10^{-3}$ &             &             &             \\
  \end{tblr}
  \label{tab:result-masksize}
  \vspace{5mm}
  \centering
  \caption{Effect of Probability of Fake Hole Occurrence}
  \begin{tblr}{
    width=0.85\linewidth,
    colsep=2mm,
    colspec={Q[l]X[c]X[c]X[c]}
    }
    \toprule
                                                                                                     & {{{$p=\SI{0.7}{\%}$}}} & {{{$p=\SI{1.4}{\%}$}}} & {{{$p=\SI{2.1}{\%}$}}} \\
    \cmidrule{1-1} \cmidrule[l]{2-4}
    Ankylosaurus                                                                                     & 1.358                  & \BB 1.317              & 1.318                  \\
    Igea                                                                                             & 2.020                  & 1.779                  & \BB 1.761              \\
    Fandisk                                                                                          & 0.900                  & \BB 0.525              & 1.173                  \\
    Part-lp                                                                                          & 1.685                  & \BB 1.353              & 4.961                  \\
    \bottomrule
    \SetCell[c=4]{c} $^*$The values in this table are $\epsilon_{\text{hole}}$ in units of $10^{-3}$ &                        &                        &                        \\
  \end{tblr}
  \label{tab:result-maskratio}
\end{table}

\begin{figure*}[t!]
  \centering
  \includegraphics[width=\linewidth]{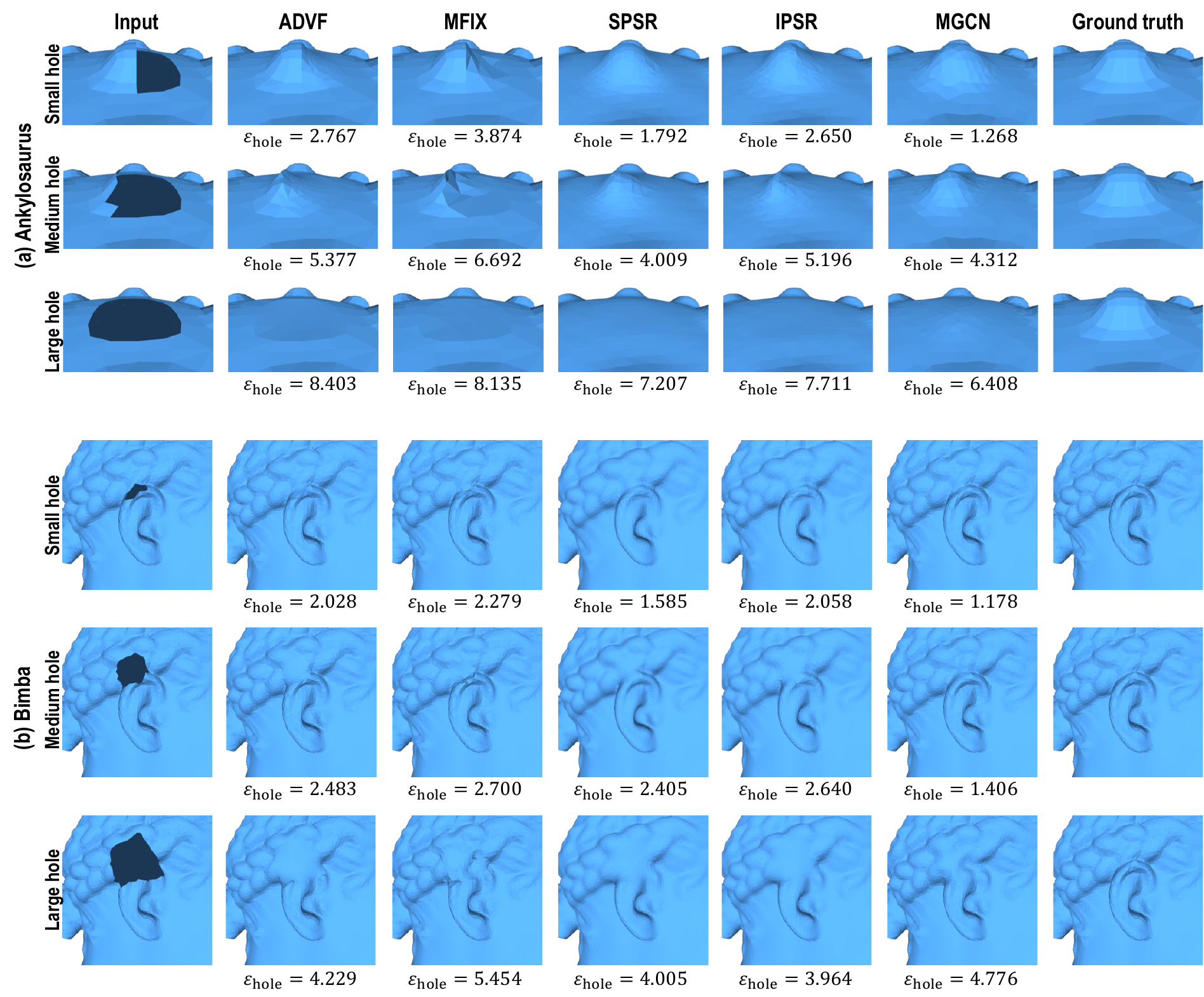}
  \caption{Mesh inpainting results for holes of different sizes, obtained by our method and prior methods.}
  \label{fig:bump-missing}
\end{figure*}

\subsection{Effect of Fake Hole Size and Occurrence}

\Cref{tab:result-masksize} shows $\epsilon_{\text{hole}}$ values obtained by MGCN trained using different mask sizes. When we select seed vertices with a fixed probability of $p=\SI{1.4}{\%}$, the mask sizes $k = 2$, $4$, and $6$ mark approximately $\SI{25}{\%}$, $\SI{60}{\%}$, and $\SI{85}{\%}$ vertices, respectively, as covered by fake holes. While the optimal mask size for each mesh varies, the results in \cref{tab:result-masksize} suggest that a relatively small mask size (e.g., $k = 2, 4$) performs better, while a large mask size (e.g., $k = 6$) may diminish the performance.

Similar behavior can be observed when varying the probability of selecting seed vertices. \Cref{tab:result-maskratio} shows $\epsilon_{\text{hole}}$ values obtained by MGCN trained with different seed occurrence probabilities. When we fix the mask size as $k=4$, using the probabilities $p= \SI{0.7}{\%}$, $\SI{1.4}{\%}$, and $\SI{2.1}{\%}$ results in marking approximately $\SI{35}{\%}$, $\SI{60}{\%}$, and $\SI{75}{\%}$ vertices, respectively, as covered by fake holes. As with mask size, lower probabilities of seed vertex occurrence often yield better scores, while a high probability of $p = \SI{2.1}{\%}$ may result in worse scores than that of lower probabilities.

Moreover, training the neural network using fewer sets of mask patterns tends to slow down the convergence. Conversely, using a larger number of mask patterns increases the time required to train the neural network over the same number of epochs. Based on these observations, we empirically determined that using 40 different mask patterns achieves a good balance between the computational efficiency and the inpainting quality.

\subsection{Performance with Varying Hole Sizes}
\label{ssec:larger-holes}

\Cref{fig:bump-missing} presents the result of performance evaluation with holes of varying sizes. Given the self-supervised learning approach of our method, it is designed to fill holes by referring to the shapes present in the non-hole regions of the input mesh. This experiment aims to explore the extent to which our method can successfully reproduce characteristic shapes when varying portions of these shapes are missing. In the case of \cref{fig:bump-missing}, we created holes in (a) a bump of the back on the ankylosaurus and (b) the hair braid pattern of the Bimba sample. Below each result, we show the positional error $\epsilon_{\text{hole}}$ calculated on the hole region.

From these results, it is evident that both our method and baseline methods can reproduce the actual geometry when the hole size is comparatively small. Notably, our method achieved the smallest $\epsilon_{\text{hole}}$ for these small holes, as can be seen in the top row of each group of results. As the hole size increases, the values of $\epsilon_{\text{hole}}$ for all the methods increase (i.e., the results get worse). Generally, our method continues to achieve the smallest values, while there are several exceptions, such as a mid-size hole in the ankylosaurus sample and a large hole in the Bimba sample. In the case of the ankylosaurus sample, all methods, including ours, failed to reproduce a bump and instead generated a smooth surface over the hole. This indicates that compensating for completely missing geometric features remains a significant challenge. In contrast, in the case of the large hole in Bimba's hair braid, our method provided the most natural braid patterns compared with baseline methods, despite not achieving the smallest $\epsilon_{\text{hole}}$.

Accordingly, although our method, as well as other baseline methods not relying on shape datasets, faces a limitation in reproducing completely missing geometric features, it still achieves the most promising results in terms of visual esthetics, even when large portions of characteristic shapes are missing.

\subsection{Limitations}
\label{ssec:limitations}

\begin{figure}[t!]
  \centering
  \includegraphics[width=\linewidth]{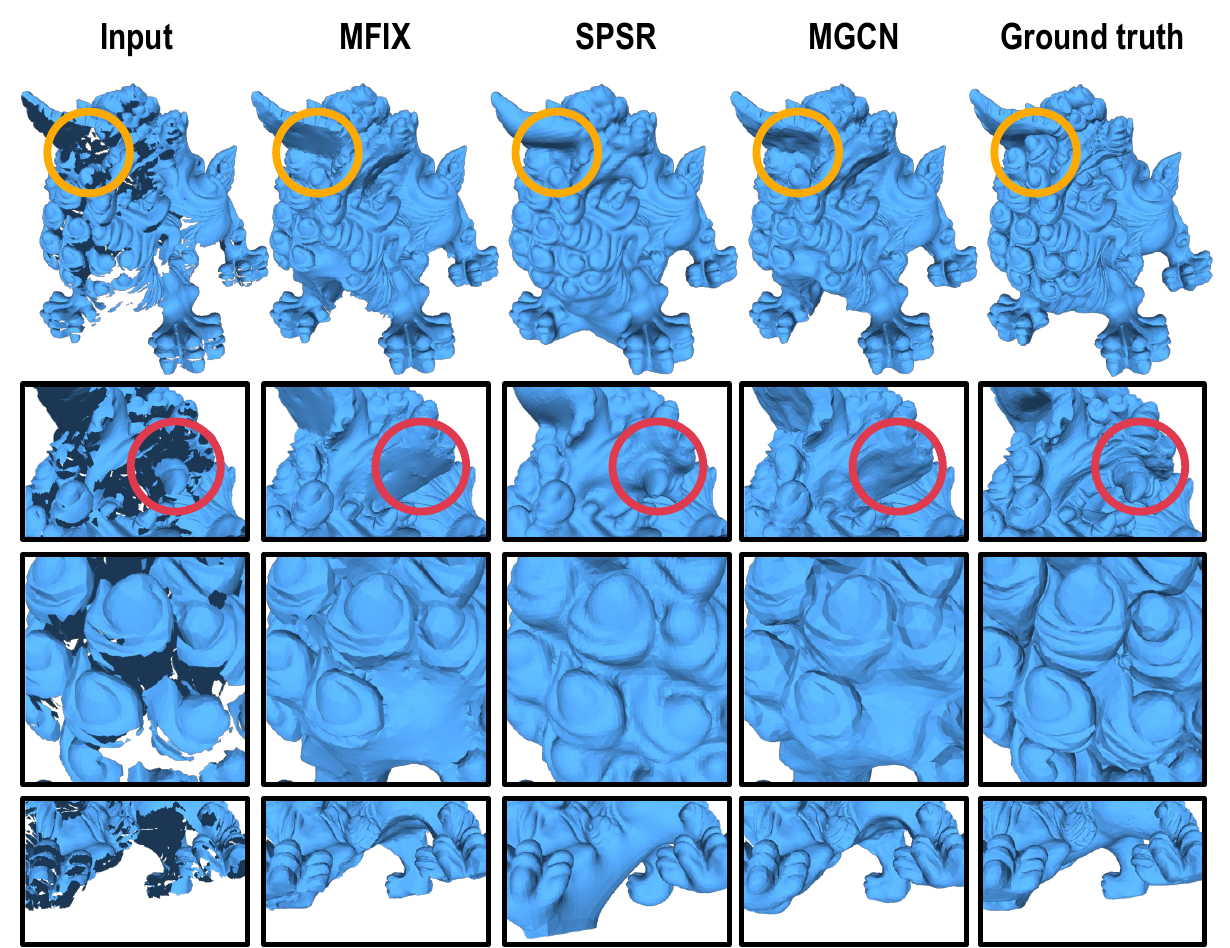}
  \caption{Failure case of our mesh inpainting where small detached portions of meshes are eliminated by MeshFix.}
  \label{fig:limitation-meshfix}
\end{figure}

While our method has shown promising results for most of the incomplete meshes we tested, it is not without limitations. One constraint is that our method requires the expected complete shape for an input incomplete mesh to be watertight and connected. This problem arises from the limitation that MeshFix retains only the largest piece of isolated mesh parts, which is demonstrated by the lion-dog example in \cref{fig:limitation-meshfix}. In this example, a part of the right fang, which is indicated by red circles, is isolated from the main body mesh and is subsequently lost during MeshFix processing. Consequently, this fang part is absent in the results of both MFIX and our MGCN, although it is retained by SPSR. To potentially overcome this problem, we could consider starting with a mesh obtained by SPSR. However, we found that this approach introduces another problem: the need to determine which parts of the input mesh should be utilized for self-supervision. The surface generated by SPSR deviates from the input mesh, even at non-hole regions, complicating the application of self-supervised learning based on the known geometries of these non-hole regions. Developing a more sophisticated initial hole-filling method that addresses these problems is an important area of future research.

Furthermore, as with previous methods that do not rely on large-scale datasets, our method faces a limitation in recovering characteristic shapes that are not observable in the non-hole regions. For example, in the lion-dog sample, both our method and baseline methods were unable to reproduce protruded crests located immediately under the right ear, as indicated by orange circles in \cref{fig:limitation-meshfix}.

Lastly, while its computation time is shorter than that of other self-prior-based methods, our method requires several minutes of training for each input. This duration is significantly longer than those of data-driven methods, such as PMP-Net~\cite{wen2021pmp}, IF-Net~\cite{chibane2020implicit}, and ShapeFormer~\cite{yan2022shapeformer}, which share a similar objective to our method but use neural networks pretrained on large-scale datasets. It is important to note that the dependence on the training datasets poses a problem in reproducing complicated shapes uncommon in such datasets. We refer the readers to the supplementary document, which shows the results for such uncommon shapes. This situation represents a trade-off, where the characteristics of self-prior-based methods and supervised-learning methods have their respective advantages and disadvantages that vary depending on the practical application. Understanding such trade-offs is important in determining the most suitable approach for a given task.

\section{Conclusion}
\label{sec:conclusion}

In this study, we investigated GCNs that acquire a self-prior for mesh inpainting by self-supervised learning. As our method leverages the self-prior, it does not require heuristics, user interaction, or large-scale training datasets. Furthermore, our method works on meshes without converting them to other shape formats, such as point clouds, voxel grids, and implicit functions. In this sense, our method can be used in the same manner as traditional approaches that are not based on machine learning, and furthermore, it can achieve better shape qualities on inpainted meshes than such traditional approaches.

In future work, we are interested in extending our method using advanced DIP variants (e.g.,~\cite{mataev2019deepred,cheng2019bayesian,jo2021rethinking}). Utilizing these sophisticated approaches may reduce the number of sets of mask patterns required for our self-supervised mesh inpainting, further decreasing computation time. Additionally, we are interested in leveraging vertex colors alongside vertex positions, as most off-the-shelf 3D scanners capture both. Lastly, although potential ideas have been proposed by recent studies~\cite{lei2021picasso,lei2021geometric,fortunato2022multiscale}, improving the time and memory efficiency of basic neural network layers (e.g., convolution and dynamic pooling/unpooling) for meshes remains an important and challenging research direction.

\ifCLASSOPTIONcompsoc
  \section*{Acknowledgments}
\else
  \section*{Acknowledgment}
\fi

We would like to thank the Stanford 3D Repository and AIM@SHAPE Shape Repository for providing the 3D models. Tatsuya Yatagawa is supported by the JSPS Grant-in-Aid for Early-Career Scientists (JP22K17907).

\ifCLASSOPTIONcaptionsoff
  \newpage
\fi



%
\printbibliography

\vfill\newpage



\clearpage


\ifsupp
  \input{supp.tex}
  \clearpage
\fi

\end{document}

%% file: supp.tex
\makeatletter
\def\maketitle{\par%
  \begingroup%
  \normalfont%
  \def\thefootnote{}
  \def\footnotemark{}
  \let\@makefnmark\relax
  \footnotesize
  \footnotesep 0.7\baselineskip
  \@IEEEcompsoconly{\long\def\@makefntext##1{\parindent 1em\noindent\hbox{\@makefnmark}##1}}\normalsize\ifCLASSOPTIONpeerreview \newpage\global\@topnum\z@ \@maketitle\@IEEEstatictitlevskip\@IEEEaftertitletext\thispagestyle{IEEEpeerreviewcoverpagestyle}\@thanks\else \if@twocolumn\ifCLASSOPTIONtechnote\newpage\global\@topnum\z@ \@maketitle\@IEEEstatictitlevskip\@IEEEaftertitletext\else \twocolumn[{\IEEEquantizevspace{\@maketitle}[\IEEEquantizedisabletitlecmds]{0pt}[-\topskip]{\baselineskip}{\@IEEENORMtitlevspace}{\@IEEEMINtitlevspace}\@IEEEaftertitletext}]\fi \else \newpage\global\@topnum\z@ \@maketitle\@IEEEstatictitlevskip\@IEEEaftertitletext\fi \thispagestyle{IEEEtitlepagestyle}\@thanks\fi
  \if@IEEEusingpubid
    \enlargethispage{-\@IEEEpubidpullup}%
  \fi
  \endgroup
  \setcounter{footnote}{0}\let\maketitle\relax\let\@maketitle\relax
  \gdef\@thanks{}%
  \let\thanks\relax}

\long\def\@IEEEtitleabstractindextextbox#1{\parbox{1\textwidth}{#1}}
\ifCLASSOPTIONcompsoc
  \long\def\@IEEEtitleabstractindextextbox#1{\parbox{0.922\textwidth}{\@IEEEcompsocnotconfonly{\rightskip\@flushglue\leftskip\z@skip}#1}} \fi

\def\@maketitle{\newpage
  \bgroup\par\vskip\IEEEtitletopspace\vskip\IEEEtitletopspaceextra\centering%
  \ifCLASSOPTIONtechnote
    {\bfseries\large\@IEEEcompsoconly{\Large\sffamily}\@title\par}\vskip 1.3em{\lineskip .5em\@IEEEcompsoconly{\large\sffamily}\@author
        \@IEEEspecialpapernotice\par}\relax
  \else
    \vskip0.2em{\Huge\ifCLASSOPTIONtransmag\bfseries\LARGE\fi\@IEEEcompsoconly{\sffamily}\@IEEEcompsocconfonly{\normalfont\normalsize\vskip 2\@IEEEnormalsizeunitybaselineskip
      \bfseries\Large}\@IEEEcompsocnotconfonly{\vskip 0.75\@IEEEnormalsizeunitybaselineskip}\@title\par}\relax
    \@IEEEcompsocnotconfonly{\vskip 0.5\@IEEEnormalsizeunitybaselineskip}\vskip1.0em\par%
    \ifCLASSOPTIONconference%
      {\@IEEEspecialpapernotice\mbox{}\vskip\@IEEEauthorblockconfadjspace%
      \mbox{}\hfill\begin{@IEEEauthorhalign}\@author\end{@IEEEauthorhalign}\hfill\mbox{}\par}\relax
    \else
      \ifCLASSOPTIONpeerreviewca
        {\@IEEEcompsoconly{\sffamily}\@IEEEspecialpapernotice\mbox{}\vskip\@IEEEauthorblockconfadjspace%
        \mbox{}\hfill\begin{@IEEEauthorhalign}\@author\end{@IEEEauthorhalign}\hfill\mbox{}\par
        {\@IEEEcompsoconly{\vskip 1.5em\relax
            \@IEEEtitleabstractindextextbox{\@IEEEtitleabstractindextext}\par\noindent\hfill
            \hfill\hbox{}\par}}}\relax
      \else
        \ifCLASSOPTIONtransmag
          {\@IEEEspecialpapernotice\mbox{}\vskip\@IEEEauthorblockconfadjspace%
          \mbox{}\hfill\begin{@IEEEauthorhalign}\@author\end{@IEEEauthorhalign}\hfill\mbox{}\par
          {\vspace{0.5\baselineskip}\relax\@IEEEtitleabstractindextextbox{\@IEEEtitleabstractindextext}\vspace{-1\baselineskip}\par}}\relax
        \else
          {\lineskip.5em\@IEEEcompsoconly{\sffamily}\sublargesize\@author\@IEEEspecialpapernotice\par
            {\@IEEEcompsoconly{\vskip 1.5em\relax
                \@IEEEtitleabstractindextextbox{\@IEEEtitleabstractindextext}\par\noindent\hfill
                \hfill\hbox{}\par}}}\relax
        \fi
      \fi
    \fi
  \fi\par\addvspace{0.5\baselineskip}\egroup}
\makeatother

\renewcommand{\thesection}{\Alph{section}}
\renewcommand{\theequation}{\Alph{section}\arabic{equation}}
\renewcommand{\thefigure}{\Alph{section}\arabic{figure}}
\renewcommand{\thetable}{\Alph{section}\arabic{table}}
\renewcommand{\thepage}{Appendix\,\arabic{page}}

\onecolumn

\setcounter{section}{0}
\setcounter{equation}{0}
\setcounter{figure}{0}
\setcounter{table}{0}
\setcounter{page}{1}

\makeatletter
\let\@oldtitle\@title
\renewcommand{\@title}{\@oldtitle\\[3mm]---Supplementary Document---}
\makeatother

\author{Shota~Hattori, Tatsuya~Yatagawa, Yutaka~Ohtake, and~Hiromasa~Suzuki}
\markboth{}{}

\IEEEtitleabstractindextext{}

\maketitle
\IEEEdisplaynontitleabstractindextext
\IEEEpeerreviewmaketitle


\section{MGCN Architecture}
\label{sec:mgcn-arch}

\paragraph{Architecture}

The multi-resolution graph convolutional network (MGCN) has an encoder--decoder architecture, but unlike the U-net architecture, skip connections are not incorporated. \Cref{fig:mgcn-arch} illustrates the MGCN architecture. As discussed in the main text, the pooling and unpooling operations that are required to define the encoder--decoder network are defined using multi-resolution meshes computed in advance by the progressive meshes method~\cite{hoppe1996progressive}.

\paragraph{Input and output}

As depicted in the figure, the input to MGCN consists of a graph structure, defined by vertex connections on the input mesh, and a set of feature vectors, each assigned to a vertex. Each 4D feature vector, mentioned earlier in the main text, is transformed by the network into a 3D vector representing the displacement needed to reproduce the completed mesh from the smoothed version of the initial mesh.

\paragraph{Multi-resolution supervision}

As demonstrated in the results of the ablation study in \cref{tab:result-loss}, achieving high-quality mesh inpainting requires multi-resolution supervision for vertex positions. This supervision is performed by forwarding the intermediate output from each UpConv block to a separate graph convolution layer (denoted as ``Conv'' in \cref{fig:mgcn-arch} and depicted by a purple arrow), which produces vertex displacements for the low-resolution version of the initial mesh.

\begin{figure*}[h!]
  \centering
  \includegraphics[width=\linewidth]{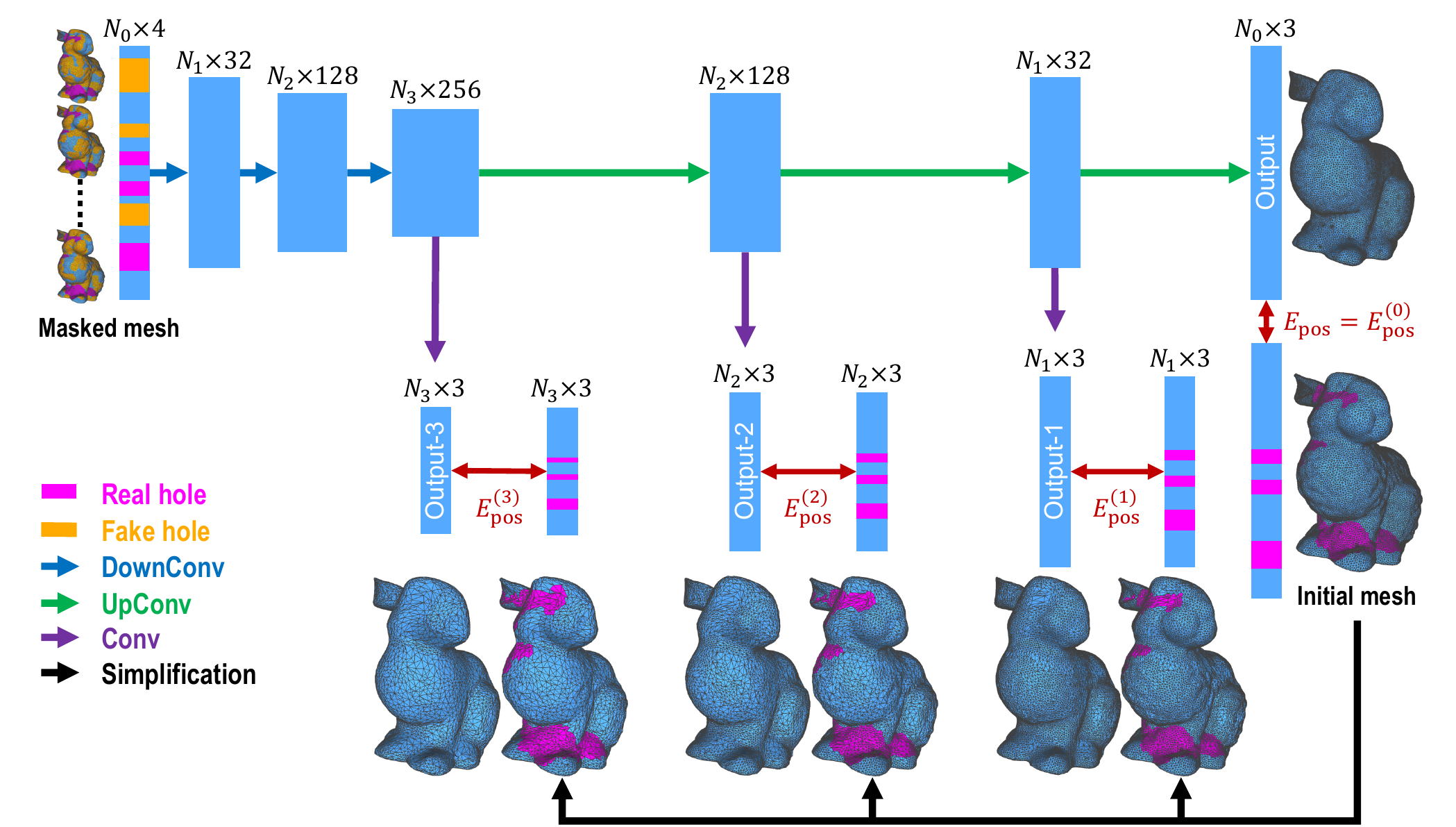}
  \caption{Illustration of the MGCN architecture.}
  \label{fig:mgcn-arch}
\end{figure*}

\clearpage

\section{Input meshes and hyperparameters}
\label{sec:input-meshes}

\Cref{fig:all-meshes} shows all the meshes with missing regions used in our experiments, and \cref{tab:all-meshes} shows the information on each mesh, i.e., the type (either ``Non-CAD'', ``CAD'', or ``Real scan''), the number of vertices and triangles, and the weight parameter $\mu$ used in the refinement step.

As for the types of meshes, the entire shape of each mesh is provided in the shape repository for the non-CAD and CAD models. We classified the models into either non-CAD or CAD ones based on whether sharp geometric features, such as edges and corners, are distinctive in each shape or not. Therefore, the non-CAD and CAD labels do not necessarily indicate whether the models were created using CAD software or not. In contrast, real-scan models shown here are prepared either using only some of the range scans provided in the shape repositories~\cite{stanford3d,aim3d} or scanning sample objects using an optical 3D scanner (GOM ATOS Core 135) by ourselves.

\vspace{10mm}

\begin{table*}[h!]
  \begin{minipage}[t]{0.5\linewidth}
    \vspace{0pt}
    \centering
    \includegraphics[width=\linewidth]{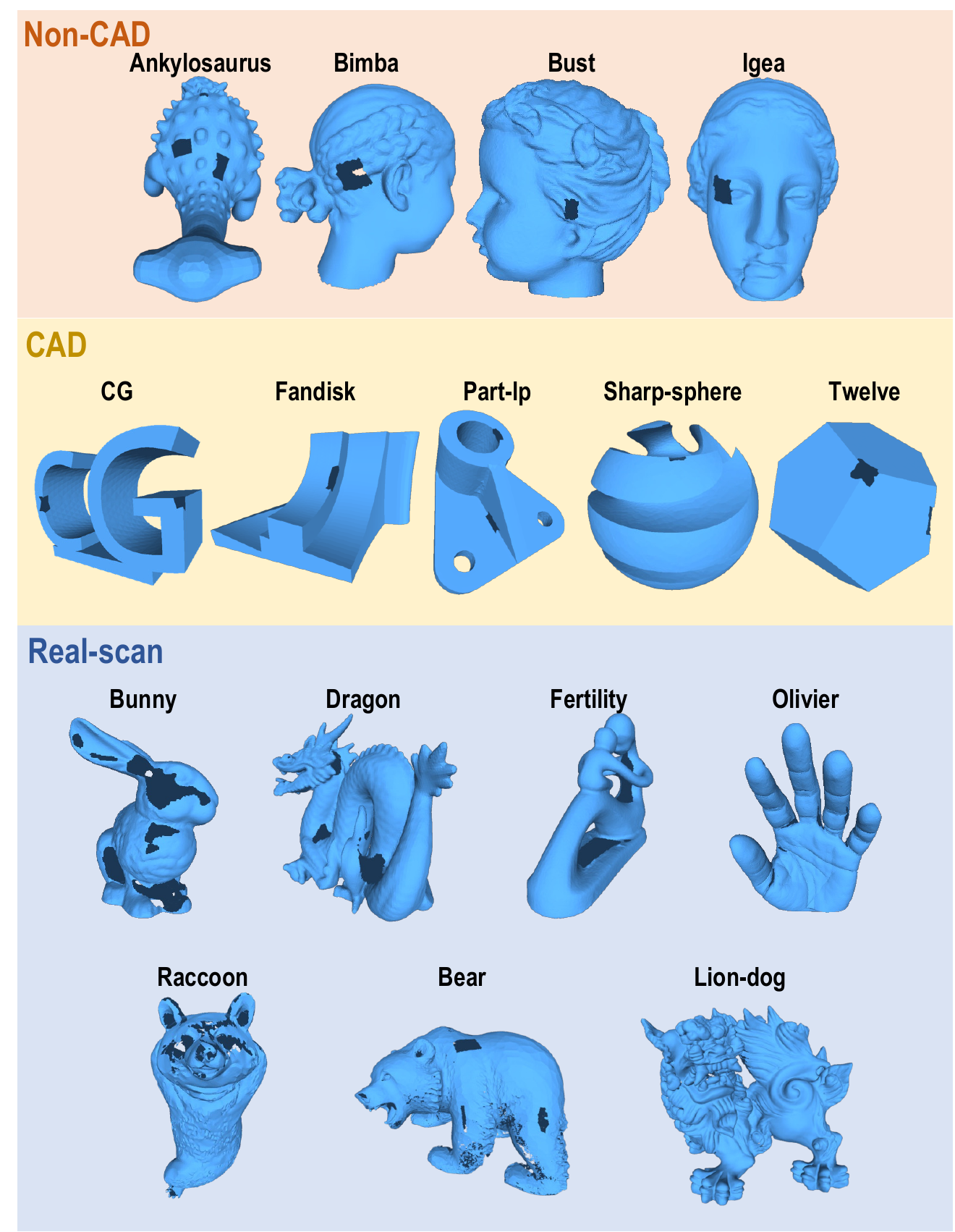}
    \captionof{figure}{All the meshes used in our experiments.}
    \label{fig:all-meshes}
  \end{minipage}
  \hfill
  \begin{minipage}[t]{0.45\linewidth}
    \vspace{0pt}\raggedright
    \centering
    \captionof{table}{Details of Meshes Used in Our Experiments}
    \label{tab:all-meshes}
    \begin{tblr}{
        colsep=3mm,
        colspec={
            Q[l]
            Q[c]
            Q[c,si={table-format=5.0}]
            Q[c,si={table-format=5.0}]
            Q[c,si={table-format=1.2}]
          }
      }
      \toprule
                   & {{{Type}}} & {{{Vertices}}} & {{{Faces}}} & {{{$\mu$}}} \\
      \cmidrule{1-1} \cmidrule[l]{2-5}
      Ankylosaurus & Non-CAD    & 21090          & 42176       & 1.0         \\
      Bimba        & Non-CAD    & 24132          & 48260       & 1.0         \\
      Bust         & Non-CAD    & 33967          & 67930       & 1.0         \\
      Igea         & Non-CAD    & 44143          & 88282       & 1.0         \\
      \cmidrule{1-1} \cmidrule[l]{2-5}
      CG           & CAD        & 9949           & 19894       & 0.1         \\
      Fandisk      & CAD        & 16724          & 33444       & 0.01        \\
      Part-lp      & CAD        & 10165          & 20338       & 0.01        \\
      Sharp-sphere & CAD        & 18084          & 36164       & 0.1         \\
      Twelve       & CAD        & 11362          & 22720       & 0.1         \\
      \cmidrule{1-1} \cmidrule[l]{2-5}
      Bunny        & Real scan  & 30272          & 60540       & 1.0         \\
      Dragon       & Real scan  & 35248          & 70496       & 1.0         \\
      Fertility    & Real scan  & 29907          & 59826       & 1.0         \\
      Olivier      & Real scan  & 29509          & 59014       & 1.0         \\
      Raccoon      & Real scan  & 30318          & 60632       & 1.0         \\
      Bear         & Real scan  & 54360          & 99999       & 1.0         \\
      Lion-dog     & Real scan  & 52077          & 99999       & 1.0         \\
      \bottomrule
    \end{tblr}
  \end{minipage}
\end{table*}

\clearpage

\section{Comparison with data-driven methods}
\label{sec:compare-data-driven}

\crefname{enumi}{}{}

In contrast to our method, which operates on an incomplete input mesh, many previous methods are based on fully-supervised deep learning. To examine the performance of such approaches on the input meshes used in our experiment, we compared our results with those of three data-driven methods: PMP-Net~\cite{wen2021pmp}, IF-Net~\cite{chibane2020implicit}, and ShapeFormer~\cite{yan2022shapeformer}. PMP-Net operates on point clouds, while the other two work on surface geometries using implicit functions as an intermediate format.

\begin{figure*}[h!]
  \centering
  \includegraphics[width=0.75\linewidth]{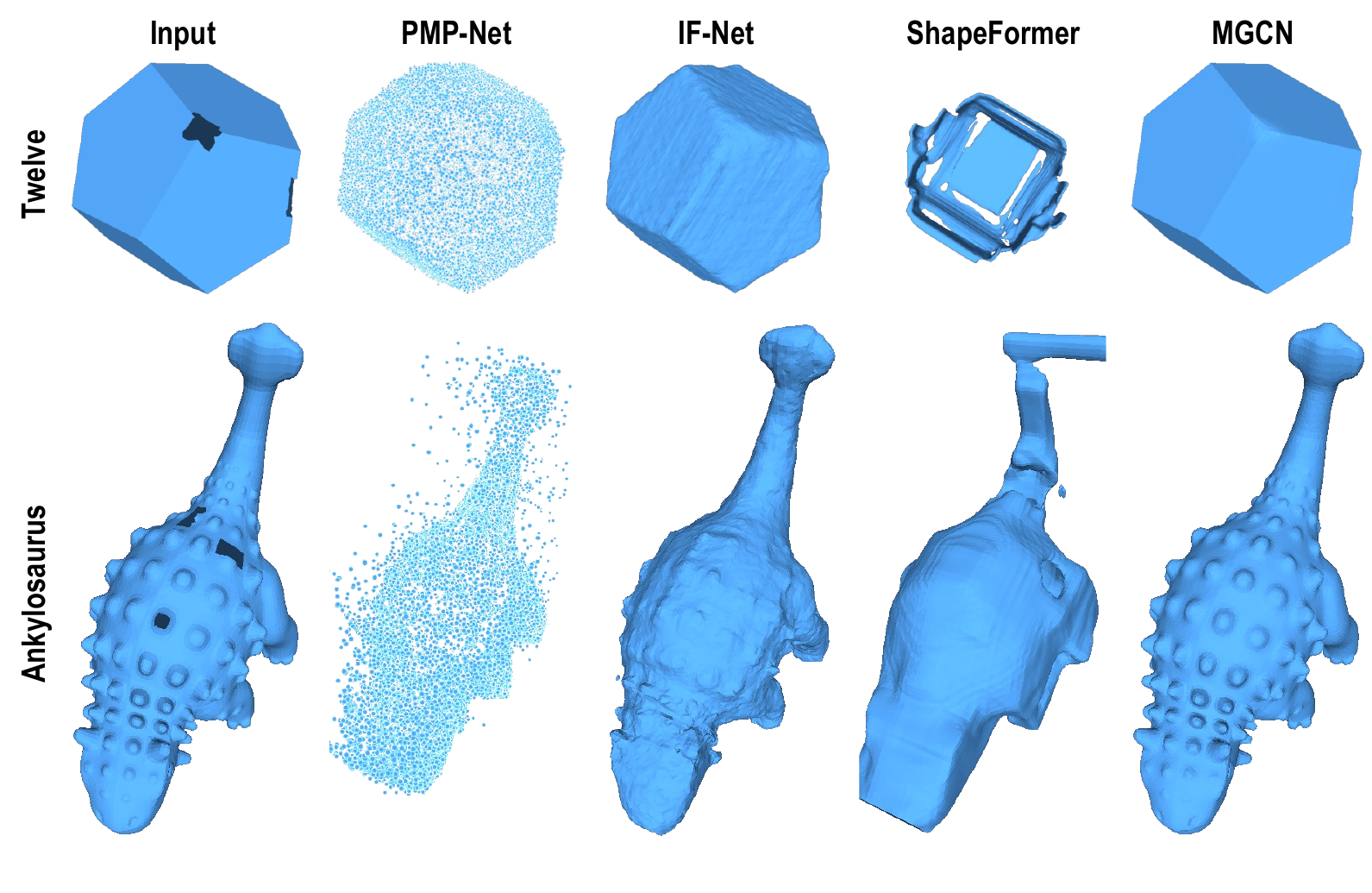}
  \caption{Results obtained by recent data-driven methods, i.e., PMP-Net~\cite{wen2021pmp}, IF-Net~\cite{chibane2020implicit}, and ShapeFormer~\cite{yan2022shapeformer}, and our method.}
  \label{fig:datadriven}
\end{figure*}

\Cref{fig:datadriven} shows the results of the data-driven methods and our method. To obtain the results in this figure, we utilized the codes and weights for pretrained neural networks provided by the authors of PMP-Net\textsuperscript{\cref{enum:pmp-net}}, IF-Net\textsuperscript{\cref{enum:if-net}}, and ShapeFormer\textsuperscript{\cref{enum:shapeformer}}. As depicted in \cref{fig:datadriven}, all these methods, unfortunately, failed to inpaint the shape geometries since these inputs (ankylosaurus and sharp sphere models in this figure) are not included in the datasets used by these data-driven methods. In contrast, our method is independent of training datasets, yet it still effectively inpaint arbitrary shapes, resulting in natural inpainted geometries.

On the other hand, data-driven methods have advantages in runtime speed. As long as a given shape is similar to some in the dataset, these methods can achieve both high quality and speed. Therefore, there is a trade-off between data-driven methods and self-prior-based approaches like our method.

\vspace{2mm}

{\footnotesize
  \begin{enumerate}[label=\arabic*.,leftmargin=5mm]
    \item PMP-Net: \url{https://github.com/diviswen/PMP-Net}, accessed on Apr. 5th, 2023 \label{enum:pmp-net}
    \item IF-Net: \url{https://github.com/jchibane/if-net}, accessed on Apr. 5th, 2023 \label{enum:if-net}
    \item ShapeFormer \url{https://github.com/qheldiv/shapeformer}, accessed on Apr. 5th, 2023 \label{enum:shapeformer}
  \end{enumerate}
}

